%% file: iclr2026_conference.tex
\documentclass{article}
\PassOptionsToPackage{table}{xcolor}
\usepackage{iclr2026_conference,times}

\input{math_commands.tex}

\usepackage{hyperref}
\usepackage{url}
\usepackage{graphicx}
\usepackage{booktabs}
\usepackage{multirow}
\usepackage{algorithm}
\usepackage{algpseudocode}
\usepackage{amssymb}
\usepackage{mathrsfs  }
\usepackage{subcaption}

\title{Learning Unified Representation of \\ 3D Gaussian Splatting}

\author{Yuelin Xin$^1$\quad Yuheng Liu$^1$\quad Xiaohui Xie$^1$\quad Xinke Li$^{2, \dagger}$\quad \\
\vspace{-0.5em}
\\
$^1$UC Irvine \quad
$^2$City University of Hong Kong\\
}

\iclrfinalcopy

\begin{document}

\renewcommand{\thefootnote}{}
\footnotetext{$^\dagger$ Corresponding author.}

\maketitle

\begin{abstract}
      A well-designed vectorized representation is crucial for the learning systems natively based on 3D Gaussian Splatting. While 3DGS enables efficient and explicit 3D reconstruction, its parameter-based representation remains hard to learn as features, especially for neural-network-based models. Directly feeding raw Gaussian parameters into learning frameworks fails to address the non-unique and heterogeneous nature of the Gaussian parameterization, yielding highly data-dependent models. This challenge motivates us to explore a more principled approach to represent 3D Gaussian Splatting in neural networks that preserves the underlying color and geometric structure while enforcing unique mapping and channel homogeneity. In this paper, we propose an embedding representation of 3DGS based on continuous submanifold fields that encapsulate the intrinsic information of Gaussian primitives, thereby benefiting the learning of 3DGS. Implementation available at \url{https://github.com/cilix-ai/gs-embedding}.
\end{abstract}

\input{sections/introduction}

\input{sections/related_works}
\input{sections/method}

\input{sections/experiments}
\input{sections/conclusion}

\clearpage
\bibliography{iclr2026_conference}
\bibliographystyle{iclr2026_conference}

\newpage
\appendix
\input{sections/appendix}

\end{document}

%% file: math_commands.tex
\usepackage{amsmath,amsfonts,bm}

\def\eqref#1{equation~\ref{#1}}

\def\1{\bm{1}}

\DeclareMathAlphabet{\mathsfit}{\encodingdefault}{\sfdefault}{m}{sl}
\SetMathAlphabet{\mathsfit}{bold}{\encodingdefault}{\sfdefault}{bx}{n}

\newtheorem{definition}{Definition}
\newtheorem{proposition}{Proposition}

%% file: sections/introduction.tex
\section{Introduction}

Recent advances in 3D Gaussian Splatting (3DGS)~\citep{kerbl20233d} have established it as a powerful technique for representing and rendering 3D scenes, enabling high-fidelity, real-time novel view synthesis through explicit parameterization of Gaussian primitives. This representation has catalyzed a growing body of work exploring learning-based methods that operate directly on Gaussian primitives, supporting tasks such as compression~\citep{shin2025locality}, generation~\citep{yi2024gaussiandreamer, xie2025generative}, and understanding~\citep{guo2024semantic}. In these pipelines, the native parameterization $\boldsymbol{\theta} = \{\boldsymbol{\mu}, \mathbf{q}, \mathbf{s}, \mathbf{c}, o\}$ is often adopted as the input or output of neural architectures.

Despite its effectiveness in optimization-based reconstruction, we identify fundamental limitations when this parametric representation is employed as a learning space for neural networks. Specifically, the native parameterization $\boldsymbol{\theta}$ conflicts with the inductive biases of standard neural architectures in three critical ways. First, the mapping from parameters to rendered output exhibits \textbf{non-uniqueness}. Ambiguities such as quaternion sign duality and symmetry-induced variances create a one-to-many mapping. This mapping creates ``embedding collisions'' where distinct parameter inputs producing identical visual outputs generate conflicting supervision signals, leading to training instability and poor convergence \citep{bengio2013representation, wang2020understanding}. Second, the parameter components suffer from \textbf{numerical heterogeneity}. Spatial positions span large magnitudes while quaternions remain unit-normalized, violating the homogeneous feature distribution assumption required for effective gradient flow \citep{ioffe2015batch}. Third, these parameters inherently reside on \textbf{distinct mathematical manifolds}, such as positions in $\mathbb{R}^3$, rotations in $SO(3)$, and appearance in spherical harmonic coefficients. Forcing these non-Euclidean variables into the Euclidean feature spaces of standard encoders breaks their intrinsic geometric structure, making the representation difficult to compress or regularize effectively.

These theoretical misalignments translate into substantial practical failures across various applications. Our empirical analysis reveals that parametric encoders are critically unstable; for instance, due to the sign ambiguity, simply negating a quaternion (a mathematically equivalent rotation) can cause complete decoding failure in parameter-trained autoencoders, see App. Fig. \ref{fig:perturb}. This instability extends to generative tasks involving latent manipulation, where we observe that linear interpolation in the parametric latent space results in geometric ``jitters'' and unnatural transitions. Furthermore, parametric embeddings lack robustness to noise, with minor perturbations causing disproportionate reconstruction errors. Crucially, for downstream tasks extensively explored in recent work including generative modeling \citep{yi2024gaussiandreamer, zhou2024diffgs}, compression \citep{shin2025locality, girish2024eagles}, and editing \citep{chen2024gaussianeditor}, these flaws manifest as discontinuous latent spaces and inefficient encoding. Instead of capturing robust geometric semantics, models are forced to resolve parameter ambiguities with more efforts, leading to sub-optimal performance in learning-based reconstruction \citep{charatan2024pixelsplat} and limited generalization across diverse domains.

We thus propose a principled alternative that represents each Gaussian primitive as a continuous field defined on its iso-probability surface. This submanifold field representation establishes a unique correspondence between Gaussians and their geometric-photometric properties, removing the ambiguities of parametric representations. By discretizing this field as a colored point cloud sampled from the probability surface, we obtain a numerically uniform and geometrically consistent representation. We further employ a variational autoencoder to learn embeddings from these discretized submanifold fields, together with a Manifold Distance metric based on optimal transport that better correlates with perceptual quality than parameter-space distances. Extensive experiments show higher reconstruction quality, stronger cross-domain generalization, and more robust latent representations. The learned embedding space is smooth and generalizable, indicating strong potential for semantic understanding and generative modeling, as demonstrated through unsupervised segmentation and neural field decoding with the proposed embeddings.
To summarize the contribution of this work, we:
\begin{itemize}
    \item  identify and formally characterize the fundamental limitations of parametric Gaussian representations for neural learning, including non-uniqueness and numerical heterogeneity.
    \item  propose a submanifold field representation that provides provably unique and geometrically consistent encoding of Gaussian primitives.
    \item develop a variational autoencoder framework incorporating a novel Manifold Distance metric based on optimal transport theory for effective learning in the submanifold field representation space with extensive experimental evidence.
\end{itemize}

\begin{figure}
    \centering
    \includegraphics[width=0.9\linewidth]{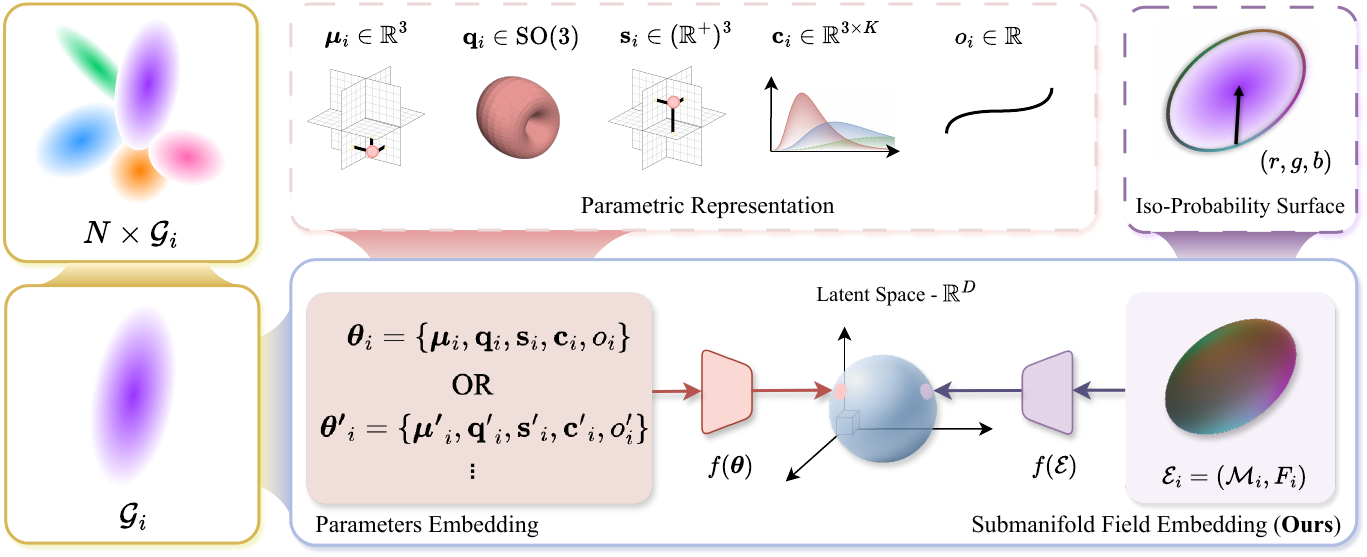}
    \vspace{-0.25cm}
    \caption{
    A scene of $N$ Gaussian primitives can be represented by $N$ sets of parameters $\boldsymbol{\theta}$ (shown in pink). Data in this parametric space resides on different manifolds and is heterogeneous and non-Euclidean, introducing challenges for encoders to fit disparate data manifolds implicitly. Shown in purple is the proposed representation, instead of relying on Gaussian parameterization, we introduce a canonical submanifold field space $(\mathcal{M}, F)$ that uniquely represents a Gaussian primitive with an iso-probability surface. 
    }
    \label{fig:parameter_manifolds}
    \vspace{-1em}
\end{figure}

%% file: sections/related_works.tex
\section{Related Works}

\textbf{3D Gaussian Splatting.} Since its re-introduction by~\cite{kerbl20233d}, 3DGS has rapidly become a core method for novel view synthesis and 3D representation. By placing explicit Gaussian primitives in 3D space and employing efficient rasterization and accumulation, 3DGS achieves real-time rendering with high fidelity~\citep{bao20253d, xinlin}. Several studies improve efficiency~\citep{jo2024identifying, lee2024gscore}, while others leverage large-scale datasets for generalization~\citep{ma2025large, li2025scenesplat}. At the application level, 3DGS has been adopted for digital human~\citep{li2024animatable, kocabas2024hugs, xiaoyuan}, self-driving scene modeling~\citep{zhou2024hugs, zhou2024drivinggaussian, yan2023street}, and physics-based simulation~\citep{jiang2024vr, xie2024physgaussian, zhong2024reconstruction}. Beyond fixed Gaussian parameters, another line augments primitives with latent embeddings to capture semantics, open-vocabulary understanding~\citep{qin2024langsplat} and deformation modeling~\citep{zhobro2025learning}. These efforts demonstrate that 3DGS provides high-fidelity appearance and serves as a versatile representation with broad potential~\citep{sun2025generalizable}.

\textbf{3DGS Parameters Regression.} To enable fast and flexible reconstruction without per-scene optimization, recent works seek to directly obtain Gaussian splats through feedforward prediction networks. For example,~\cite{charatan2024pixelsplat, chen2024mvsplat} proposed to predict Gaussian parameters directly from multi-view input, while~\cite{zheng2024gps} generate pixel-wise parameter maps and lift them to 3D via depth estimation. This paradigm has been extended to pose-free settings~\citep{hong2024pf3plat, chen2024pref3r, tian2025drivingforward}, and transformer-based methods further improve scalability and generalization~\citep{li2025vicasplat, jiang2025anysplat, lin2025longsplat}.
Overall, these methods directly output Gaussian parameters from neural networks, showing that 3DGS can serve as an effective target for network-based prediction in efficient reconstruction.

\textbf{Embedding Gaussian Primitives.}
Recent works move beyond reconstruction and encode Gaussian parameters into latent spaces for tasks such as generation, editing, and compression.
~\cite{zhou2024diffgs, lin2025diffsplat, wewer2024latentsplat} learn structured latent variables from 3D Gaussian space to fulfill generation tasks. 
Editing and style-transfer methods use diffusion or style conditioning to manipulate Gaussians primitives in latent or rendering spaces~\citep{chen2024gaussianeditor, igs2gs, lee2025editsplat, palandra2024gsedit, zhang2024stylizedgs, kovacs2024, yu2024instantstylegaussian}.
Other works improve rendering quality by optimizing Gaussian parameters under diffusion priors~\citep{tang2023dreamgaussian, yi2024gaussiandreamer, chen2024text}, while compression methods~\citep{girish2024eagles,yang2025hybridgs} reduce storage and computation by quantizing and embedding Gaussian parameters.
These approaches show the potential of embedding Gaussians into neural latent spaces, but they assume Gaussian parameters are naturally compatible with neural learning, overlooking that these parameters were designed for optimization-based reconstruction. This oversight underlies our analysis in Section~\ref{sec:method} and our proposal of a more suitable formulation.

%% file: sections/method.tex
\section{Method}
\label{sec:method}

\subsection{Preliminaries: Gaussian Splatting Parameterization}

A scene under 3D Gaussian Splatting is represented as a set of $N$ oriented, and view-dependently colored Gaussian primitives $\{\mathcal{G}_i\}_{i=1}^N$, each contributing to the final rendered image via rasterization and alpha compositing. Each Gaussian primitive $\mathcal{G}_i$ is usually represented by a parameter tuple $\boldsymbol{\theta}_i = \{\boldsymbol{\mu}_i, \mathbf{q}_i, \mathbf{s}_i, \mathbf{c}_i, o_i\}$, where:

\begin{itemize}
    \item $\boldsymbol{\mu}_i \in \mathbb{R}^3$: the center position of the Gaussian in world coordinates;
    \item $\mathbf{q}_i \in \text{SO}(3)$: a unit quaternion representing the local rotation;
    \item $\mathbf{s}_i \in (\mathbb{R}^{+})^{3}$: scale parameters along the rotated axes;
    \item $\mathbf{c}_i \in \mathbb{R}^{3 \times K}$: spherical harmonic (SH) coefficients for view-dependent color for $K \in \mathbb{Z}$;
    \item $o_i \in \mathbb{R}$: a logit-transformed opacity value $\alpha_i = \sigma(o_i)$, where $ \sigma$ is a sigmoid function.
\end{itemize}

The local geometry of the Gaussian is governed by its covariance matrix, constructed as
\begin{equation}
\label{eq:sigma}
\Sigma_i = R(\mathbf{q}_i) \, \text{diag}(\mathbf{s}_i)^2 \, R(\mathbf{q}_i)^\top,
\end{equation}
where $R(\mathbf{q}_i)$ is the rotation matrix corresponding to the quaternion $\mathbf{q}_i$. This defines an ellipsoidal spatial density, centered at $\boldsymbol{\mu}_i$, whose shape and orientation determine the contribution of $\mathcal{G}_i$ to the rendered scene. The color at a given view direction $\mathbf{d} \in \mathbb{S}^2$ is computed per channel using SH basis functions denoted by
\begin{equation}\label{eq:sh_formula}
\text{Color}_i(\mathbf{d}) = \left[\text{SH}^r_i(\mathbf{d}),\text{SH}^g_i(\mathbf{d}),\text{SH}^b_i(\mathbf{d})\right]^{\top},
\end{equation}
where $\text{SH}^c_i(\mathbf{d})$ in $c-$channel is calculated by $\sum_{l=0}^{L_{\text{max}}} \sum_{m=-l}^{l} (\mathbf{c}_i)_{c,(l,m)} \cdot Y_l^m(\mathbf{d})$ $Y_l^m$ is the real-valued spherical harmonic of degree $l$ and order $m$. The final rendering aggregates contributions from all $\mathcal{G}_i$ via a soft visibility-weighted compositing process. This native parameterization is well-suited for gradient-based scene optimization. However, it introduces significant challenges when used as a representation for learning.

\subsection{Parameterization is Ill-Suited as a Learning Space}

The parameter representation $\boldsymbol{\theta}$ poses fundamental challenges when used as a learning space for neural networks. We identify two critical issues: representation non-uniqueness and numerical heterogeneity. Each undermines the stability and effectiveness of neural network training.

\textbf{Representation Non-uniqueness.} The parametric representation suffers from a many-to-one mapping that violates basic requirements for stable learning. To understand this, we first formalize what rendering effect a single Gaussian primitive produces. 
\begin{definition}[Single Gaussian Radiance Field (SGRF)]
A SGRF is a radiance field $\phi: \mathbb{R}^3 \times \mathbb{S}^2 \rightarrow {\mathbb{R}}^3$, 
The field is defined by the local density at point $\mathbf{x}\in\mathbb{R}^3$ along direction $\mathbf{d}\in\mathbb{S}^2$:
\begin{equation}
\phi_{\mathcal{G}}(\mathbf{x},\mathbf{d}) = \rho_{\mathcal{G}}(\mathbf{x}) \cdot c_{\mathcal{G}}(\mathbf{d}),
\end{equation}
where $\rho_{\mathcal{G}}(\mathbf{x}) = \exp\left(-\frac{1}{2}(\mathbf{x}-\boldsymbol{\mu})^{\top} \Sigma^{-1}(\mathbf{x}-\boldsymbol{\mu})\right)$ is a volume density function and $c_{\mathcal{G}}(\mathbf{d})$ is a color radiance field coupled with opacity.  Specifically, given  a parameter set  $\boldsymbol{\theta} = \{\boldsymbol{\mu}, \mathbf{q}, \mathbf{s}, \mathbf{c}, o\}$, $\Sigma$ can be derived by ~\eqref{eq:sigma} and $c_{\mathcal{G}}(\mathbf{d}) =\sigma(o)\cdot \operatorname{Color}(\mathbf{d})$ can be derived by ~\eqref{eq:sh_formula}.
\end{definition}

The SGRF, derived from the multi-Gaussian rendering framework by \cite{kerbl20233d}, specifies how the final value at any pixel is rendered in a scene containing only one Gaussian splat. Furthermore, let $\Phi$ be the space of SGRFs, and $\Theta \subseteq \mathbb{R}^{|\boldsymbol{\theta}|} $ be the paramater space of Gaussian primitives, each parameter set $\boldsymbol{\theta}\in \Theta$ provides a complete representation that generates a correponding field $\phi_{\mathcal{G}} \in \Phi$.  We indicate that a single SGRF may correspond to multiple parameterizations of Gaussian primitives, as formalized in the following proposition.

\begin{proposition}[Non-uniqueness of the SGRF Parametric Representation]
The parametric representation of a SGRF is not unique. Formally, there exist at least two distinct parameter sets, 
$\boldsymbol{\theta}_1 \in \Theta $ and $\boldsymbol{\theta}_2 \in \Theta $ with $\boldsymbol{\theta}_1 \neq \boldsymbol{\theta}_2$, that generate the exact same field $\phi_{\mathcal{G}}\in \Phi$.

\end{proposition}

The non-uniqueness is from quaternion sign ambiguity, geometric symmetries, and rotation-spherical harmonic interactions producing equivalent parameter combinations (see App. \ref{app:prop_1_proof} for proof). The non-uniqueness of $\boldsymbol{\theta}$ will create ``embedding collisions'' where different parameter vectors produce identical rendered output~\citep{wang2020understanding}. This makes the learning objective $\|\boldsymbol{\theta}_\text{pred} - \boldsymbol{\theta}_\text{target}\|_p$ ambiguous, as multiple parameter configurations can achieve the same visual result. The resulting conflicting gradients lead to training instability and poor convergence indicated by \cite{bengio2013representation}.

\textbf{Numerical Heterogeneity} The parameter components violate the homogeneous distribution assumption of standard neural architectures. Neural networks typically assume features share similar statistical properties for effective gradient flow \citep{ioffe2015batch}. However, 3D Gaussian parameters span vastly different ranges. For example, pre-activation scales can range from $-15$ to $3$, while quaternions stay unit-normalized. More fundamentally, these parameters follow different distributions and live on different manifolds: positions $\boldsymbol{\mu} \in \mathbb{R}^3$, rotations $\mathbf{q} \in \text{SO}(3)$, scales $\mathbf{s} \in (\mathbb{R}^+)^3$, and SH coefficients $c$ with exponential decay. Concatenating them ignores their heterogeneous nature. Small noises in quaternions can drastically alter geometry, while small noise in high-order SH coefficients is negligible, yet the network treats all dimensions equally.

The non-uniqueness and numerical heterogeneity of the native parameter space $\boldsymbol{\theta}$ make it unsuitable for neural network learning, which would generate unstable embeddings (see our experiments in Sec. \ref{sec:latent} and App. \ref{app:extra_results}). We therefore introduce a submanifold field representation that ensures unique mappings and respects the geometric structure of 3D Gaussians.

\subsection{Representation on  Submanifold Field}

To address this issue, we propose converting each Gaussian primitive $\mathcal{G}_i$ to a novel geometric representation $\mathcal{E}_i$, which is a color field defined on a 2D submanifold in 3D Euclidean space, as illustrated in Fig.~\ref{fig:parameter_manifolds}. For a Gaussian density $\mathcal{N}(\mathbf{x};\boldsymbol{\mu}_i,\Sigma_i)$, we define the iso-probability surface at fixed radius $r$ as:
\begin{equation}\label{eq:manifold}
\mathcal{M}_i = \left\{\mathbf{x}\in\mathbb{R}^3 \mid (\mathbf{x}-\boldsymbol{\mu}_i)^\top\Sigma_i^{-1}(\mathbf{x}-\boldsymbol{\mu}_i) = r^2\right\},
\end{equation}
which forms an ellipsoid surface, namely, a two-dimensional submanifold, centered at $\boldsymbol{\mu}_i$. On this  submanifold, we define a field function:
\begin{equation}
F_i(\mathbf{x}) = \sigma(o_i)\cdot\text{Color}_i(\mathbf{d}_\mathbf{x}),
\end{equation}
where $\mathbf{d}_\mathbf{x} = (\mathbf{x}-\boldsymbol{\mu}_i)/\|\mathbf{x}-\boldsymbol{\mu}_i\|$ denotes the unit direction vector for $\mathbf{x}\in\mathcal{M}_i$, and $\text{Color}_i(\cdot)$ represents the view-dependent color parameterization as in~\eqref{eq:sh_formula}. Let $\mathbb{M}$ be the space of all possible iso-probability submanifolds as defined in ~\eqref{eq:manifold}, we define our unified representation space as:
\begin{equation}
 \mathscr{E} = \left\{\mathcal{E}_i=(\mathcal{M}_i, F_i) \mid \mathcal{M}_i \in \mathbb{M}, \, F_i: \mathcal{M}_i \rightarrow \mathbb{R}^3\right\},
\end{equation}
The representation $\mathcal{E}_i\in \mathscr{E}$ encodes both geometric properties (shape, orientation) via $\mathcal{M}_i$ and appearance attributes (view-dependent color) via $F_i$ in a continuous framework. We have the following proposition (proof is provided in App. \ref{app:injective_proof}).

\begin{proposition}[Uniqueness of Submanifold Field Representation]
For every SGRF $\phi_{\mathcal{G}}\in \Phi$, there exists a unique corresponding representation  $\mathcal{E}\in \mathscr{E}$. This establishes a one-to-one correspondence between the elements of $\Phi$ and $\mathscr{E}$ . Formally, for any two distinct fields $\phi_{\mathcal{G},1}, \phi_{\mathcal{G},2} \in \Phi$, their corresponding representations $\mathcal{E}_1, \mathcal{E}_2 \in\mathscr{E}$ are also distinct.
\end{proposition}

The  submanifold field \(\mathcal{E}_i\) thus provides a numerically stable and provably unique representation space on which we can safely build learning objectives and neural architectures.

\subsection{Encode Submanifold Fields as Embeddings}
\label{sec:autoencoder}

\begin{figure}
    \centering
    \includegraphics[width=\linewidth]{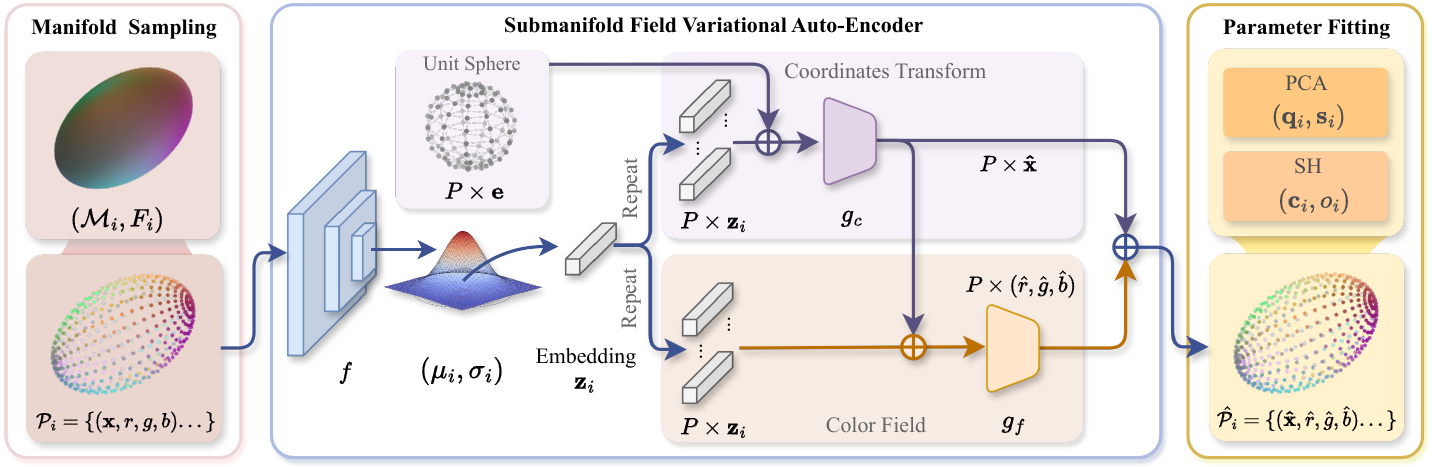}
    \vspace{-0.5cm}
    \caption{To embed the proposed submanifold field representation into a vector form suitable for neural networks, we devise a Submanifold Field Variational Auto-encoder (SF-VAE) that embeds any input submanifold field as a 32-D vector, then reconstructs the original parameter set $\boldsymbol{\theta}_i$. SF-VAE learns in our new representation space instead of the parametric space.}
    \label{fig:embedding_model}
    \vspace{-1em}
\end{figure}

We design a variational auto-encoder to encode submanifold field representation, shown in Fig. \ref{fig:embedding_model}. The network architecture, learning objectives and dataset are introduced. 

\textbf{Encoder-decoder Architecture.} We employ a point-cloud-based network to encode and decode one sub-manifold field. Particularly, we uniformly sample \(P\) points from the submanifold field $\mathcal{E}$ as a colored point cloud \(\mathcal{P}=\bigl\{(\mathbf{x}_{m},F(\mathbf{d}_{\mathbf{x}_{m}}))\bigr\}_{m=1}^{P}\). 
We then employ a PointNet~\citep{qi2017pointnet} encoder $f$ to obtain latent embedding by $\mathbf{z} \sim f(\mathbf{z} \mid \mathcal{P})$
where $\mathbf{z}\in \mathbb{R}^D$ is the embedding with dimension $D$. The decoder $g$ consists of two neural networks, namely, the coordinates transform network $g_c:\mathbb{R}^3\times \mathbb{R}^D \rightarrow \mathbb{R}^3$ and color field $g_f:\mathbb{R}^3\times \mathbb{R}^D \rightarrow \mathbb{R}^3$. The decoded point cloud from decoder is given by,
\begin{equation}
\hat{\mathcal{P}} = g(\mathbf{z}, \mathcal{U}_{P'} )=\{g_c([\mathbf{e}_n,\mathbf{z}]),g_f(\left[g_c([\mathbf{e}_n,\mathbf{z}]),\mathbf{z}\right])\}_{n=1}^{P'}
\end{equation}
where $\mathcal{U}_{P'}=\{\mathbf{e}_n\}_{n=1}^{P'}$ is a set of coordinates sampled from a unit sphere surface. Such canonical set works as the initial input for two implicit functions $g_c$ and $g_f$, and queries new coordinates and color field.
Furthermore, to recover the original Gaussian parameters $\boldsymbol{\theta}_i$ for rendering purposes, we estimate the covariance matrix $\Sigma_i$ by principal component analysis (PCA), and SH coefficients $\mathbf{c}_i$ by fitting the spherical harmonics to $\hat{\mathcal{P}} $.

\textbf{Learning Objectives.} We introduce \textbf{\emph{Manifold Distance}} (M-Dist) for the reconstruction objective in encoder-decoder training. Given two submanifold fields $\mathcal{E} = (\mathcal{M}, F)$ and $\hat{\mathcal{E}} = (\hat{\mathcal{M}}, \hat{F})$, we propose to measure their similarity based on the Wasserstein-2 distance from optimal transport
defined as
\begin{equation}
W_2^2(\mathcal{E}, \hat{\mathcal{E}}) = \inf_{\gamma \in \Gamma(\hat{\sigma}, \hat{\sigma}')} \int_{\mathcal{M} \times \hat{\mathcal{M}}} d^2\big((\mathbf{x}, 
c_x), (\mathbf{y},c_y
)\big) d\gamma(\mathbf{x}, \mathbf{y})
\end{equation}
where  $c_x = F(\mathbf{d}_{\mathbf{x}})$, $c_y = \hat{F}(\mathbf{d}_{\mathbf{y}})$, $\Gamma(\hat{\sigma}_i, \hat{\sigma}_j)$ is the set of all joint probability measures (transport plans) with marginals $\hat{\sigma}_i$ and $\hat{\sigma}_j$, and the ground distance is defined as
\begin{equation}
d^2\big((\mathbf{x}, c_x), (\mathbf{y}, c_y)\big) = \|\mathbf{x} - \mathbf{y}\|_2^2 + \lambda \|c_x - c_y\|_2^2,
\end{equation}
with $\lambda \in \mathbb{R}^+$ balancing spatial and color terms.  In practice, both $\mathcal{M}$ and $\hat{\mathcal{M}}$ are discretized as colored point clouds $\mathcal{P}$ and $\hat{\mathcal{P}}$. The empirical Wasserstein-2 distance $\hat{W}$ is then computed between these point clouds by
\begin{equation}
\hat{W}_2^2(\mathcal{P}, \hat{\mathcal{P}}) = \min_{\mathbf{\Gamma} \in \Gamma(\hat{\sigma}, \hat{\sigma}')} \sum_{(\mathbf{x}_i, c_{x_i}) \in \mathcal{P}} \sum_{(\mathbf{y}_j, c_{y_j}) \in \hat{\mathcal{P}}} \mathbf{\Gamma}_{ij} \left(d^2\big((\mathbf{x}_i, c_{x_i}), (\mathbf{y}_j, c_{y_j})\big)\right).
\end{equation}
Finally, the learning objective for variational auto-encoder is
\begin{equation}
\mathcal{L}_{\text{VAE}} =\mathbb{E}_{\hat{\mathcal{P}}\sim \operatorname{VAE}(\mathcal{P})}\left(\hat{W}_2^2(\mathcal{P}, \hat{\mathcal{P}} ) + \beta \cdot d_{\text{KL}}\left(f(\mathbf{z}\mid \mathcal{P})\|\mathcal{N}(0,\mathbf{I})\right)\right),
\end{equation}
where $\operatorname{VAE}(\mathcal{P})=g(f(\mathbf{z}\mid\mathcal{P}), \mathcal{U}_{P'})$ and the second term is the KL divergence loss for variational auto-encoder implementation, and $\beta$ is a balance factor.

\textbf{Dataset Preparation.} Since this embedding model only encodes single Gaussian primitives, which have no semantic meaning out of a scene's global context, we can use a \emph{randomly generated dataset} of Gaussian primitives to train this model, thus making it domain-invariant to data. The implementation details of this generated dataset can be found in App. \ref{app:implementation}.

%% file: sections/experiments.tex
\section{Experiments}

\subsection{Evaluation Setup}
\label{sec:exp_setup}

\textbf{Baseline Implementation Details.} 
To isolate the effect of representation choice, we adopt a self-implemented encoder–decoder framework for both the parametric representation $\boldsymbol{\theta}$ and the proposed submanifold field representation $\mathcal{E}$. While comparisons with existing 3DGS learning methods are possible, they typically involve task-specific architectures that confound the role of representation itself. Direct reuse of prior pipelines would not yield a controlled comparison, so we implement both baselines in the same VAE-style framework to attribute differences solely to the representation.

We implement and train three size-matched embedding models: our submanifold field VAE (Sec.~\ref{sec:autoencoder}), and two baseline parametric VAEs operating directly on $\boldsymbol{\theta}$. For the parametric models, each Gaussian primitive is represented as a 56-D vector ($3{+}4{+}3K{+}1$ for $L_\mathrm{max}{=}3$), omitting global coordinates to match the SF-VAE setting. A three-layer MLP encodes this input to a 32-D latent ($56\!\!\rightarrow\!\!512\!\!\rightarrow\!\!512\!\!\rightarrow\!\!32\times2$), and the decoder, either uses a MLP to map the latent back to $\widehat{\boldsymbol{\theta}}_i$, or uses the same decoder of SF-VAE to map to $\hat{\mathcal{P}}$. This setting further decouples evaluation results with the training objective functions. Apart from input dimension, all models share identical depth, width, latent size (32), and optimizer settings (all using Adam), ensuring a matched capacity.

\textbf{Datasets.} We evaluate the proposed representation and compare it with the baseline primarily using two datasets. For object-level tasks, we utilize ShapeSplat \citep{ma2025large}, a large-scale 3DGS dataset derived from ShapeNet \citep{shapenet2015}, comprising 52K objects across 55 categories. For scene-level experiments, we employ Mip-NeRF 360 \citep{barron2022mipnerf360unboundedantialiased}, which contains 7 medium-scale scenes with abundant high-frequency details. Additionally, unless stated otherwise, we train the embedding models using the randomly generated Gaussian primitive dataset, with 500K randomly generated data samples; implementation details are provided in App. \ref{app:implementation}.

\textbf{Evaluation Metrics.} To comprehensively assess both perceptual fidelity and representation quality, we report PSNR, SSIM, and LPIPS on rasterized reconstructions against ground truth Gaussian splats, as well as $L_1$ distance in the Gaussian parameter space. Crucially, we also include our proposed Manifold Distance (M-Dist) as an evaluation criterion. By cross-comparing M-Dist with parameter-space distances ($L_1$/$L_2$), we can demonstrate that M-Dist aligns more closely with perceptual “gold standard” metrics such as PSNR and LPIPS, validating our claims in Sec. \ref{sec:method}

\subsection{ Evaluation on Representation Learning Framework}

\begin{table}[t]
\centering
\caption{Reconstruction quality comparison for object-level (ShapeSplat) and scene-level (Mip-NeRF 360) datasets. All models trained on the randomly generated dataset. The three models have a parameter count of 0.62M, 0.66M and 0.62M respectively. The relatively extreme perceptual metrics values in ShapeSplat come from the use of background during measurement.}
{\footnotesize
\begin{tabular}{lll|ccc|cc}
\toprule
\textbf{Input Representation} & \textbf{Encoder} & \textbf{Decoder} & \textbf{PSNR} $\uparrow$ & \textbf{SSIM} $\uparrow$ & \textbf{LPIPS} $\!\downarrow$ & \textbf{M-Dist} & $L_1$-Dist \\
\hline
\rowcolor{gray!10}
\multicolumn{8}{c}{\textbf{ShapeSplat}} \\
\hline
Parametric & MLP & MLP & 37.512 & 0.888 & 0.152 & 0.184 & \underline{0.040} \\
Parametric & MLP & SF-VAE & 44.725 & 0.896 & 0.136 & 0.051 & 0.097 \\
Submanifold Field & SF-VAE & SF-VAE & \textbf{63.408} & \textbf{0.990} & \textbf{0.010} & \underline{0.041} & 0.098 \\
\hline
\rowcolor{gray!10}
\multicolumn{8}{c}{\textbf{Mip-NeRF 360}} \\
\hline
Parametric & MLP & MLP & 18.818 & 0.564 & 0.452 & 0.510 & \underline{0.034} \\
Parametric & MLP & SF-VAE & 20.923 & 0.730 & 0.359 & 0.055 & 0.173 \\
Submanifold Field & SF-VAE & SF-VAE & \textbf{29.833} & \textbf{0.953} & \textbf{0.079} & \underline{0.048} & 0.179 \\
\bottomrule
\end{tabular}
}
\label{tab:reconstruction}
\vspace{-1px}
\end{table}

\begin{figure}[t]
    \centering
    \includegraphics[width=\linewidth]{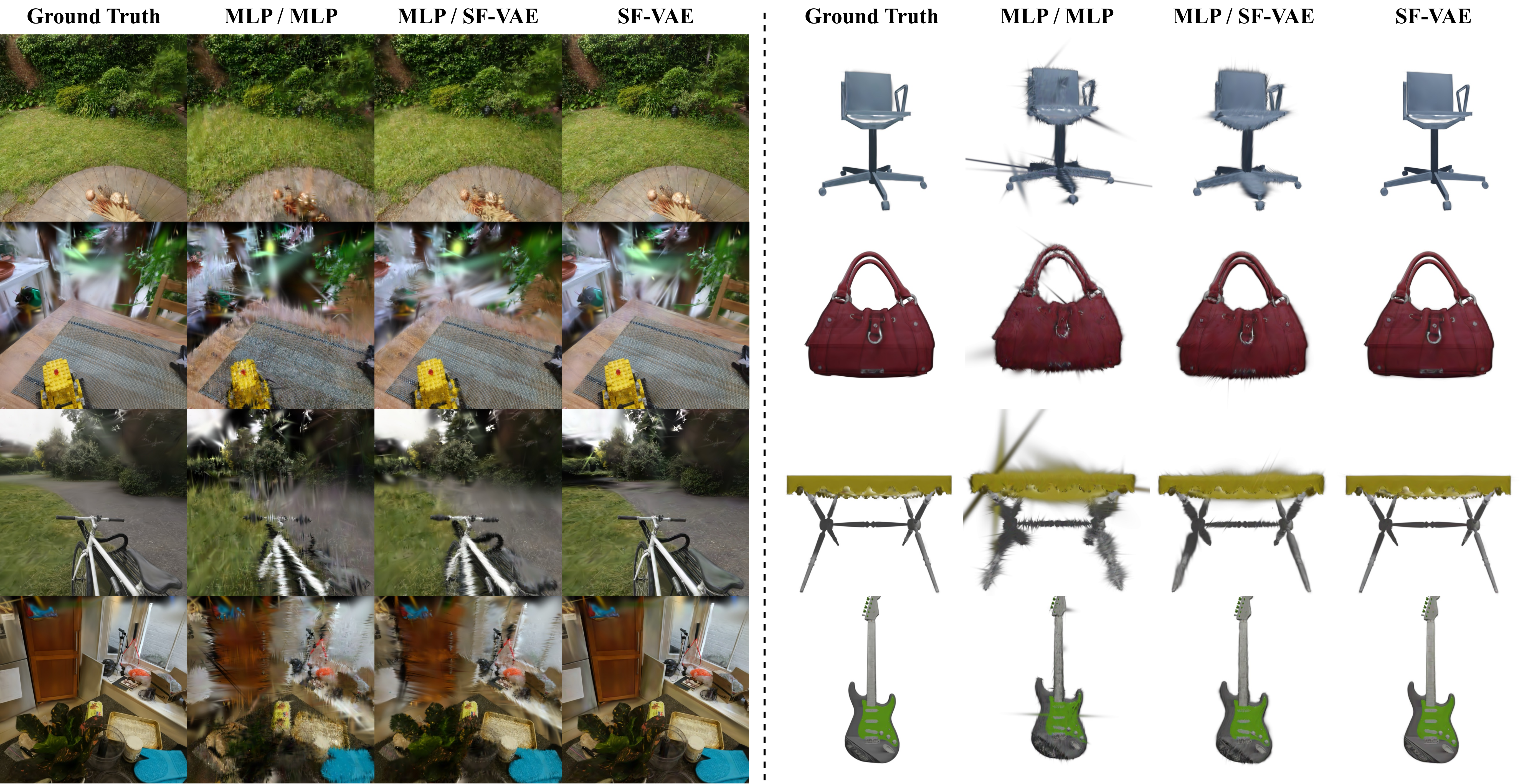}
    \vspace{-0.5cm}
    \caption{Qualitative results for rasterized reconstruction. Samples selected arbitrarily from Mip-NeRF 360 and ShapeSplat. Parametric models can induce confusion in parameter space, failing to embed and restore the correct Gaussian parameters.}
    \label{fig:render_result}
    \vspace{-1em}
\end{figure}

\textbf{Zero-shot Reconstruction.} We present a comprehensive quantitative and qualitative analysis of reconstruction quality for both object-level and scene-level data, as summarized in Tab.~\ref{tab:reconstruction} and Fig.~\ref{fig:render_result}. All models are trained on the same randomly generated 3D Gaussian primitives dataset and evaluated on ShapeSplat and Mip-NeRF 360, using three matched encoder-decoder configurations to control for bias. Across all perceptual metrics (PSNR, SSIM, LPIPS), the submanifold field representation consistently outperforms parametric baselines. For example, on ShapeSplat, SF-VAE achieves substantially higher PSNR and SSIM and a much lower LPIPS, indicating both improved fidelity and perceptual quality. Similar performance gains are observed in scene-level reconstruction, where the submanifold field model demonstrates better performance across diverse spatial contexts.

Importantly, the \textit{Manifold Distance} (M-Dist) metric shows a stronger empirical correlation with quality metrics like PSNR and LPIPS than traditional $L_1$ parameter distances, supporting our claim that M-Dist is a more robust and meaningful similarity measure for 3D Gaussian representations, truthfully reflecting perceptual differences rather than merely parameter discrepancies. The consistent improvement margin across both datasets highlights the advantage of learning in the submanifold field space, which better preserves intrinsic structure and view-dependent appearance, confirming the efficacy of our representation for high-fidelity 3D Gaussian modeling.

\begin{table}[t]
\centering
{\footnotesize
\caption{\textcolor{black}{Reconstruction quality comparison under cross-domain setting. All models trained on either ShapeSplat or Mip-NeRF 360 dataset are tested on another dataset. We show that the generalization ability of SF Embedding framework is inherently domain-agnostic even without random data.}}
\label{tab:generalization}
\begin{tabular}{ll|lc|ccc}
\toprule
\textbf{Train set} & \textbf{Test set} & \textbf{Input Represent.} & \textbf{Encoder / Decoder} & \textbf{PSNR} $\uparrow$ & \textbf{SSIM} $\uparrow$ & \textbf{LPIPS} $\downarrow$ \\
\midrule
\multirow{3}{*}{ShapeSplat} & \multirow{3}{*}{Mip-NeRF.} 
    & Parametric & MLP / MLP & 9.753 & 0.356 & 0.615 \\
    & & Parametric & MLP / SF-VAE & 14.845 & 0.675 & 0.336 \\
    & & Submanifold Field & SF-VAE / SF-VAE & 19.194 & 0.821 & 0.309 \\
\midrule
\multirow{3}{*}{Mip-NeRF.} & \multirow{3}{*}{ShapeSplat} 
    & Parametric & MLP / MLP & 55.624 & 0.957 & 0.067 \\
    & & Parametric & MLP / SF-VAE & 60.777 & 0.987 & 0.013 \\
    & & Submanifold Field & SF-VAE / SF-VAE & 62.576 & 0.990 & 0.014 \\
\bottomrule
\end{tabular}
}
\vspace{-0.5em}
\end{table}

\textbf{Cross-domain Reconstruction.} We also evaluate cross-domain generalization by training on one real-world dataset and testing on the other (object-level $\leftrightarrow$ scene-level) under an identical training protocol and capacity budget as in the reconstruction study. Concretely, we train either the proposed SF-VAE or the parametric MLP baselines on a source set and evaluate on a target set, rendering novel views and reporting PSNR, SSIM, and LPIPS averaged over test samples (see Tab. ~\ref{tab:generalization}). 
Across both transfer directions, the SF-based embedding consistently achieves higher reconstruction quality than the parametric baseline, indicating reduced sensitivity to domain-specific statistics (e.g., scale, lighting, SH complexity). Particularly, comparing these transfer results with Tab.~\ref{tab:reconstruction}, we find that the model trained on synthetic random data actually outperforms the models transferred from real-world domains. This indicates that our random learning strategy effectively strips away domain priors, which establishes our approach as a unified representation, where a single model trained on synthetic data can fundamentally generalize to real-world contexts.

\subsection{Sensitivity Study of Representation}
\label{sec:latent}

\begin{figure}[t]
    \centering
    \begin{minipage}[t]{0.75\linewidth}
        \centering
        \includegraphics[width=\linewidth]{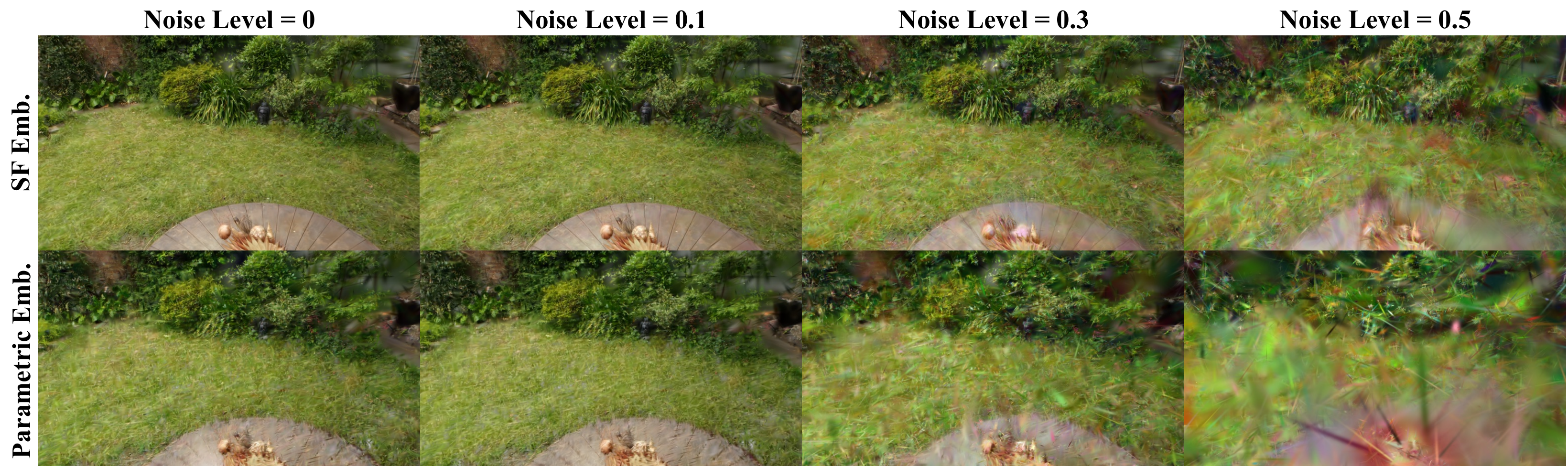}
    \end{minipage}
    \hfill
    \begin{minipage}[t]{0.24\linewidth}
        \centering
        \includegraphics[width=\linewidth]{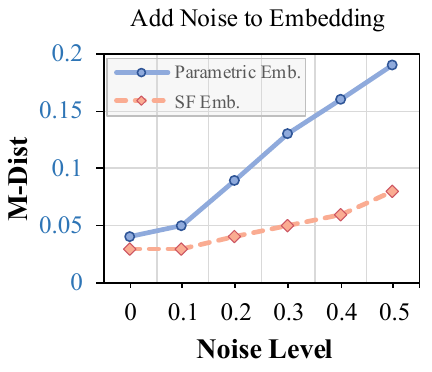}
    \end{minipage}
    \vspace{-0.52cm}
    \caption{\textcolor{black}{Reconstruction results using embeddings with noise. \textbf{Left:} Visualization of reconstructed scene from noisy embeddings of Gaussian parameters (MLP) and SF-VAE. \textbf{Right:} Comparison on M-Dist for different noise levels added to embedding space, tested on Mip-NeRF 360. \textit{Noise level} is defined as the ratio between the noise magnitude and the embedding variance.}}
    \label{fig:emb_noise}
    \vspace{-1em}
\end{figure}

\textbf{Robustness to Noise.} 
To evaluate the submanifold field embedding's robustness to noise, we gradually add higher levels of gaussian noise to the embedding space of the parametric model and the submanifold field model and test their reconstruction quality and M-Dist. To ensure fair comparison, we use noise level as a ratio to variance instead of absolute noise magnitude. As shown in Fig. \ref{fig:emb_noise}, the embedding space of submanifold field model is more robust to random perturbation, this makes submanifold field embeddings a better learning target since it is less sensitive to potential noise introduced by downstream regression.

\textbf{Latent Space Interpolation.} 
To evaluate the regularity of the latent space of the proposed representation, we randomly sample pairs of source and target Gaussian primitives $\mathcal{G}_s$ and $\mathcal{G}_t$ and linearly interpolate each pair for a fixed number of steps $n=7$. 
Compared with parametric space, the interpolation in submanifold field embedding space shows a smooth transition path, while interpolation in parametric space shows undesired jitter in rotation and scale, indicating space irregularities, see App. \ref{app:extra_results}. This highlights the motivation to learn in the unified submanifold field embedding space.

\subsection{Representation Applicability}

\begin{figure}[t]
    \centering
    \includegraphics[width=\linewidth]{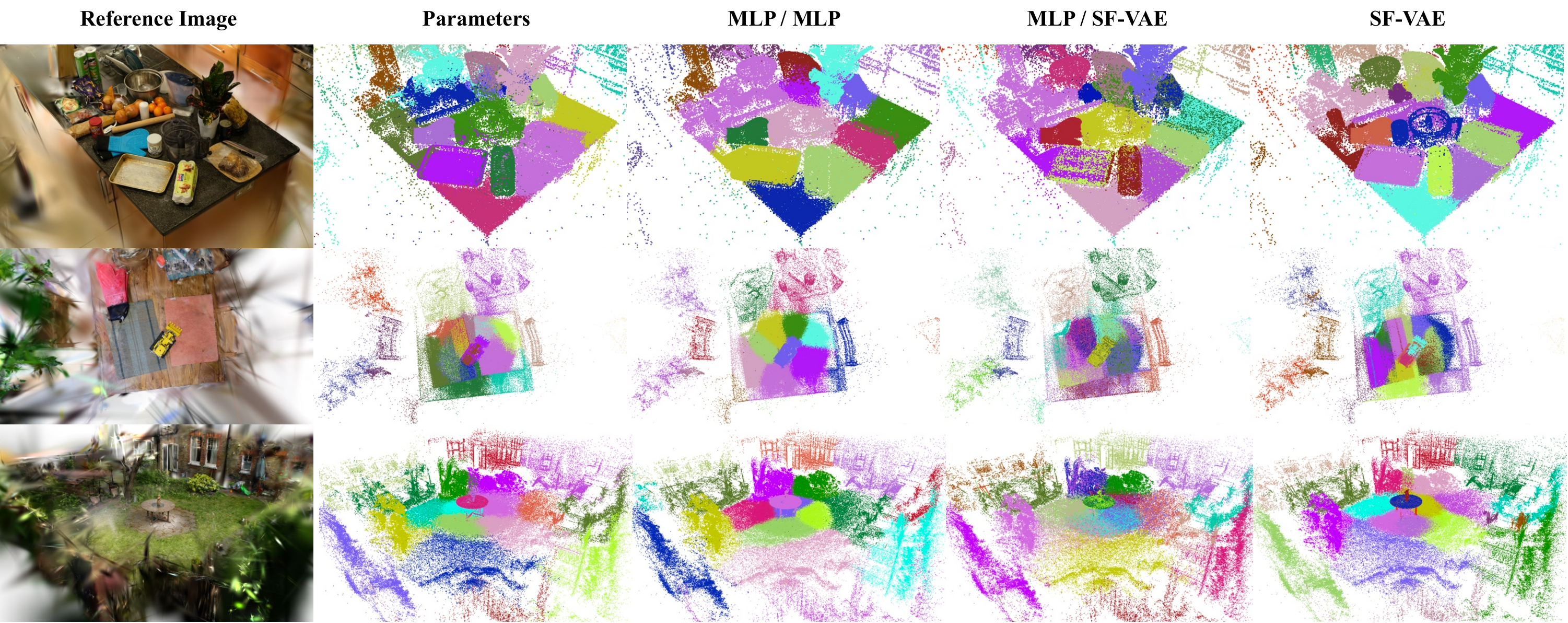}
    \vspace{-0.58cm}
    \caption{\textcolor{black}{Unsupervised graph clustering based on raw Gaussian parameters and various  embeddings. Submanifold field embeddings show better preservation of detailed semantics, showing its downstream applicability.}}
    \label{fig:cluster}
    \vspace{-1.2em}
\end{figure}

\textbf{Unsupervised Clustering.} 
To further probe the semantic structure of the learned embedding spaces, we perform unsupervised graph clustering on both the raw Gaussian parameter space and the embedding outputs of each model. As visualized in Fig.~\ref{fig:cluster}, clusters formed in the submanifold field embedding space exhibit more detailed semantic separation against the reference images compared to those formed using normalized parameters or parametric embeddings. For example, SF-VAE's embedding clustering in the first line of Fig. \ref{fig:cluster} outlines clearer separation of foreground objects with the background. The clusters appear smoother, less noisy, and with clearer boundaries, showing an ability to distinguish between different entities. This indicates that the submanifold field embedding captures more dense semantics and discriminative features, validating its usefulness.

\begin{figure}[t]
\centering
\begin{minipage}[t]{0.32\linewidth}
    \centering
    \vspace{0pt}
    \includegraphics[width=0.9\linewidth]{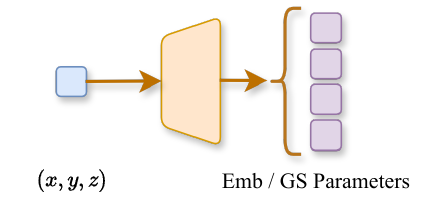}
    \vspace{-0.5em}
    \caption{Setting of a Gaussian Neural Field, we compare between the prediction target SF embedding and raw GS parameters.}
    \label{fig:genf}
\end{minipage}
\hfill
\begin{minipage}[t]{0.64\linewidth}
    \centering
    \vspace{0pt}
    \footnotesize
    \captionof{table}{\textcolor{black}{Comparison between Gaussian Neural Fields trained using submanifold field embeddings and raw Gaussian parameters. Top: ShapeSplat, bottom: Mip-NeRF 360.}}
    \label{tab:genf}

    \begin{tabular}{l|ccc|c}
        \toprule

        \textbf{Target} & \textbf{PSNR} $\uparrow$ & \textbf{SSIM} $\uparrow$ & \textbf{LPIPS} $\downarrow$ & \# Params\\
        \midrule
        Raw GS Parameter & 51.660 & 0.925 & 0.141 & 0.21M \\
        SF Embedding & 58.619 & 0.980 & 0.043 & 0.20M \\
        \midrule
        Raw GS Parameter  & 19.922 & 0.648 & 0.410 & 1.87M \\
        SF Embedding & 24.395 & 0.804 & 0.261 & 1.85M \\
        \bottomrule
    \end{tabular}
\end{minipage}
\vspace{-1em}
\end{figure}

\textbf{Gaussian Neural Fields.} To validate the potential of our representation for advanced downstream tasks, we introduce the \emph{Gaussian Neural Field} (GNF). Drawing inspiration from the decoding structures in generative diffusion models (e.g., DiffGS by \cite{zhou2024diffgs}) and neural compression frameworks \citep{wu2024implicit}, the GNF functions as a coordinate-based neural implicit field as illustrated in Fig. \ref{fig:genf}. Specifically, it employs a lightweight MLP (architecture detailed in App. \ref{app:gnf}) to learn a continuous mapping from spatial coordinates $\mathbf{x}_i$ to per-primitive descriptors. This setup allows us to evaluate the ``learnability'' of our representation: while regressing heterogeneous raw parameters $\boldsymbol{\theta}_i$ often leads to optimization difficulties, our unified SF embeddings provide a smooth and well-conditioned target for the neural field. As evidenced in Tab.~\ref{tab:genf} and visualization in App. \ref{app:gnf}, the SF-guided GNF outperforms the parameter-based baseline in visual fidelity with equivalent training effort. This indicates that our representation is more friendly to neural networks, hinting at its utility for potential downstream generative and compression tasks.

\subsection{More Studies on Implementation Details}

\begin{figure}[t]
    \centering
    \includegraphics[width=\linewidth]{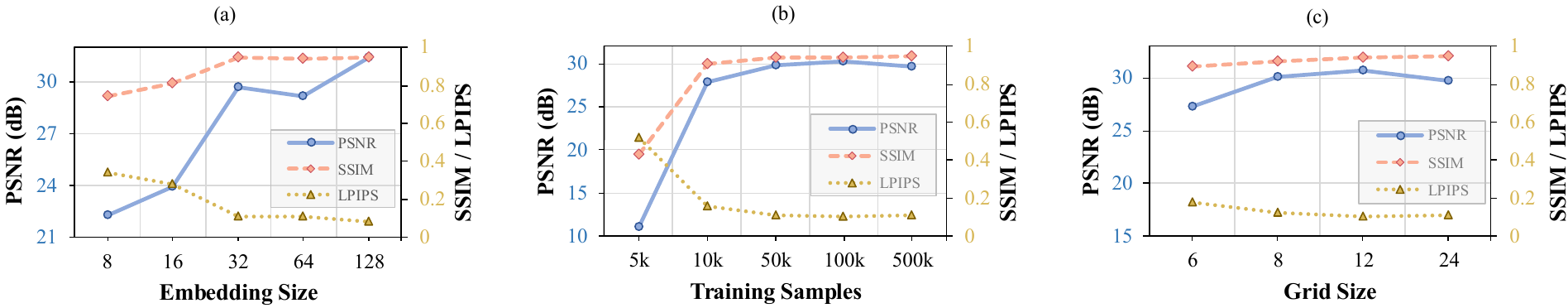}
    \vspace{-2em}
    \caption{Behavior studies tested on Mip-NeRF 360. From left to right: (a) embedding dimension, (b) generated training dataset size, \textcolor{black}{(c) Submanifold Field discretized (i.e., point sample) grid size.}}
    \label{fig:ablation}
    \vspace{-1.5em}
\end{figure}

\textbf{Ablation Study on SF-VAE Designs.}
We provide performance comparison with different framework designs based on Mip-NeRF 360. (1) For encoder $f$, we tested DGCNN encoder~\citep{dgcnn} beyond our implementation, where the DGCNN encoder achieves comparable reconstruction fidelity while it is roughly 1.75$\times$ slower in encoding and uses roughly 2$\times$ more GPU memory during inference; (2) For the decoder's unit sphere grid, we tested a 2D grid implementation with matching grid size, 2D grid achieves similar reconstruction results but takes more iterations to converge; (3) We implemented two versions of the fitting module: a GPU-based version using FP32 with batching and Cholesky decomposition, and a CPU-based one using FP64 without batching and Least Squares solver. Experiments show that the GPU version introduces only negligible quality degradation (0.4 PSNR and 0.01 SSIM), while achieving an average speedup of 85$\times$ with a batch size of 4096.

\textbf{Behavior Study of Latent Space Dimension.}
We evaluated different embedding space dimensions for the SF-VAE model to meet the best trade-off between compression and reconstruction quality. All models are trained on the generated dataset with $L=3$ order Spherical Harmonics. Results shown in Fig. \ref{fig:ablation} (a), 32 is the optimal balance point between reconstruction quality and latent space compression. All values are tested with baseline input/output of $P=12^2$. While this work does not specifically focus on compression effectiveness, embedding space robustness shown in Sec. \ref{sec:latent} suggests potential in further latent tokenization and quantization.

\textbf{Behavior Study of Training Set Size.}
To determine the number of random training samples required to achieve the best reconstruction results, we vary the trainiing sample size from 5K to 500K (baseline), see Fig. \ref{fig:ablation} (b). The results indicate the proposed representation is data efficient. When only using 2\% of the baseline training sample, our model can achieve close-to-baseline performance.

\textbf{Behavior Study of SF Discretized Size.}
To ensure the submanifold fields are truthfully represented in a discrete manner, we evaluate different sample sizes $P$, see Fig. \ref{fig:ablation} (c). Going from the lowest tested $P=6^2$ to the baseline $P=12^2$, we observe a gradual improvement in reconstruction quality, while going above $P=12^2$ yields very little improvement. Since $P$ directly correlates to the computational efficiency of the submanifold field model (see below), we keep $P=12^2$.

\label{sec:comp_eff}

\textbf{Computational Efficiency.} 
Increasing the point sample size $P$ increases computation and memory. Encoding time remains low since a lightweight PointNet-based encoder shares weights with all input points, giving an inference speed of 1.72s per 1 million Gaussians for $P=12^2$ with a batch size of 4096 on an RTX 5090. Decoding time is 4.20s per 1 million Gaussians, from embedding to Gaussian parameters. The complexity is $O(P)$ or $O(n^2)$ w.r.t. the grid size $n$. We utilize the advantage of GPU parallel computation to boost the calculation for parameter fitting module. In the 4.20s of decoding time, the fitting module (PCA + SH fitting) consumes only about 0.48s which is negligible for large Gaussian scenes.

%% file: sections/conclusion.tex
\section{Conclusion and Limitations}

We introduced a geometry-aware \emph{submanifold field} representation for 3D Gaussian Splatting that maps each primitive to a color field on a canonical iso–probability ellipsoid and proved the mapping is injective over core attributes. Built on this representation, our SF–VAE learns semantically meaningful latents and yields higher-fidelity reconstructions and stronger zero-shot generalization than capacity-matched raw-parameter baselines; our manifold distance (M-Dist) further aligns training and evaluation with geometric/perceptual similarity. 

\textbf{Limitations and Outlook.} Our current setup operates at the single-Gaussian level, while this ensures data invariance, it omits explicit inter-splat structure modeling for more complex representation learning. Promising directions include set/scene-level encoders with permutation-invariant attention, point cloud to 3DGS inpainting, generative modeling with submanifold field embeddings, temporal extensions to dynamic scenes, and applications to compression, retrieval, and regularization in broader downstream 3DGS pipelines.

%% file: sections/appendix.tex
\section{Proof of Proposition 1}
\label{app:prop_1_proof}

We construct two distinct parameter sets, $\boldsymbol{\theta}_1 = \{\mathbf{q}_1, \mathbf{s}_1, \mathbf{c}_1, o_1\}$ and $\boldsymbol{\theta}_2 = \{\mathbf{q}_2, \mathbf{s}_2, \mathbf{c}_2, o_2\}$ with $\boldsymbol{\theta}_1 \neq \boldsymbol{\theta}_2$, that generate the identical Single Gaussian Radiance Field (SGRF). The identity $L_{\mathcal{G}_1}(\mathbf{x}, \mathbf{d}) = L_{\mathcal{G}_2}(\mathbf{x}, \mathbf{d})$ for all $(\mathbf{x}, \mathbf{d})$ requires two conditions to be met:

\begin{enumerate}
    \item \textbf{Geometric Equivalence}: The covariance matrices must be equal, $\boldsymbol{\Sigma}_1 = \boldsymbol{\Sigma}_2$, which implies that the volume densities $\sigma_{\mathcal{G}}(\mathbf{x})$ are identical. We also assume equal opacity, $o_1=o_2$.
    \item \textbf{Appearance Equivalence}: The view-dependent color functions must be identical, which requires $\text{SH}(\mathbf{c}_1, \mathbf{R}_1^\top \mathbf{d}) = \text{SH}(\mathbf{c}_2, \mathbf{R}_2^\top \mathbf{d})$ for all $\mathbf{d} \in \mathbb{S}^2$, where $\mathbf{R}$ is the rotation matrix for the quaternion $\mathbf{q}$.
\end{enumerate}

We construct a distinct parameter set $\boldsymbol{\theta}_2$ by considering a discrete symmetry of the Gaussian ellipsoid. Let $\boldsymbol{\theta}_1 = \{\mathbf{q}_1, \mathbf{s}_1, \mathbf{c}_1, o_1\}$ be an initial parameterization.

Let $\mathbf{R}_{\text{flip}}$ be a rotation matrix corresponding to a 180-degree rotation about one of the local axes (e.g., the z-axis), such that $\mathbf{R}_{\text{flip}} = \text{diag}(-1, -1, 1)$. $\mathbf{R}_{\text{flip}}$ is its own inverse, $\mathbf{R}_{\text{flip}}^\top = \mathbf{R}_{\text{flip}}$. We define a new parameter set $\boldsymbol{\theta}_2$ as follows:
\begin{itemize}
    \item Let the new rotation be $\mathbf{R}_2 = \mathbf{R}_1 \mathbf{R}_{\text{flip}}$. This defines a new quaternion $\mathbf{q}_2 \neq \mathbf{q}_1$.
    \item Let the scales and opacity remain unchanged: $\mathbf{s}_2 = \mathbf{s}_1$ and $o_2 = o_1$.
\end{itemize}

First, we verify geometric equivalence. The new covariance matrix $\boldsymbol{\Sigma}_2$ is:
\begin{align*}
\boldsymbol{\Sigma}_2 &= \mathbf{R}_2 \text{diag}(\mathbf{s}_2)^2 \mathbf{R}_2^\top = (\mathbf{R}_1 \mathbf{R}_{\text{flip}}) \text{diag}(\mathbf{s}_1)^2 (\mathbf{R}_1 \mathbf{R}_{\text{flip}})^\top \\
&= \mathbf{R}_1 \left( \mathbf{R}_{\text{flip}} \text{diag}(\mathbf{s}_1)^2 \mathbf{R}_{\text{flip}}^\top \right) \mathbf{R}_1^\top
\end{align*}
Since $\mathbf{R}_{\text{flip}}$ is diagonal, it commutes with the diagonal scaling matrix $\text{diag}(\mathbf{s}_1)^2$, meaning the term in parentheses equals $\text{diag}(\mathbf{s}_1)^2$. Therefore,
\[
\boldsymbol{\Sigma}_2 = \mathbf{R}_1 \text{diag}(\mathbf{s}_1)^2 \mathbf{R}_1^\top = \boldsymbol{\Sigma}_1.
\]
Geometric equivalence is satisfied.

Next, for appearance equivalence, we must find SH coefficients $\mathbf{c}_2$ such that:
\[
\text{SH}(\mathbf{c}_1, \mathbf{R}_1^\top \mathbf{d}) = \text{SH}(\mathbf{c}_2, (\mathbf{R}_1 \mathbf{R}_{\text{flip}})^\top \mathbf{d}) = \text{SH}(\mathbf{c}_2, \mathbf{R}_{\text{flip}}^\top \mathbf{R}_1^\top \mathbf{d}).
\]
Let $\mathbf{v} = \mathbf{R}_1^\top \mathbf{d}$ be the view direction in the local frame of the first Gaussian. The condition becomes $\text{SH}(\mathbf{c}_1, \mathbf{v}) = \text{SH}(\mathbf{c}_2, \mathbf{R}_{\text{flip}}^\top \mathbf{v})$.
This states that the function defined by $\mathbf{c}_2$ when evaluated on a transformed vector must equal the function defined by $\mathbf{c}_1$ on the original vector. This is equivalent to stating that the function itself has been rotated. The properties of spherical harmonics guarantee that for any rotation, there exists a linear transformation (the Wigner D-matrix $\mathbf{D}$) that maps the original coefficients to the new ones. We can therefore find $\mathbf{c}_2$ such that:
\[
\mathbf{c}_2 = \mathbf{D}(\mathbf{R}_{\text{flip}}) \mathbf{c}_1.
\]

We have constructed a parameter set $\boldsymbol{\theta}_2 = \{\mathbf{q}_2, \mathbf{s}_1, \mathbf{c}_2, o_1\}$, which is distinct from $\boldsymbol{\theta}_1$ (since $\mathbf{q}_2 \neq \mathbf{q}_1$) yet defines the identical radiance field. This discrete symmetry exists for any Gaussian, proving the general proposition.

\par\noindent
Furthermore, this non-uniqueness expands from a discrete set to a continuous manifold of solutions for symmetric geometries.
\begin{itemize}
    \item \textbf{Case A (Isotropic)}: If $\mathbf{s}=(s,s,s)^\top$, the covariance matrix $\boldsymbol{\Sigma}=s^2\mathbf{I}$ is invariant under \textit{any} rotation. This gives rise to a continuous, multi-parameter family of equivalent solutions.
    \item \textbf{Case B (Spheroidal)}: If two scale components are equal (e.g., $\mathbf{s}=(s_a, s_a, s_b)^\top$), $\boldsymbol{\Sigma}$ is invariant to any rotation around the local axis of symmetry, resulting in a one-parameter continuous family of redundant solutions.
\end{itemize}
In all such cases, a corresponding transformation on the SH coefficients preserves appearance equivalence. Since non-unique parameterizations exist in all cases, the proposition is proven.

\section{Proof of Uniqueness of the Representation $\mathcal{E}_i = (\mathcal{M}_i, F_i)$}
\label{app:injective_proof}
Assume $\phi_{\mathcal{G}_1} = \phi_{\mathcal{G}_2}$, namely, $\rho_{\mathcal{G}_1}(\mathbf{x})=\rho_{\mathcal{G}_2}(\mathbf{x}), \forall \mathbf{x}\in \mathbb{R}^3$ and $c_{\mathcal{G}_1}(\mathbf{d})=c_{\mathcal{G}_2}(\mathbf{d}), \forall \mathbf{d}\in \mathbb{S}^2$.
We show this leads to $\mathcal{E}_1 = \mathcal{E}_2$ for two SGRFs. This implies both $\mathcal{M}_1 = \mathcal{M}_2$ and $F_1 = F_2$.

The volume densities are equal:
$$
\exp\left(-\frac{1}{2}(\mathbf{x} - \boldsymbol{\mu}_1)^\top \mathbf{\Sigma}_1^{-1} (\mathbf{x} - \boldsymbol{\mu}_1)\right) = \exp\left(-\frac{1}{2}(\mathbf{x} - \boldsymbol{\mu}_2)^\top \mathbf{\Sigma}_2^{-1} (\mathbf{x} - \boldsymbol{\mu}_2)\right)
$$
Taking the natural logarithm of both sides yields that the quadratic forms are identical for all $\mathbf{x}$:
$$
(\mathbf{x} - \boldsymbol{\mu}_1)^\top \mathbf{\Sigma}_1^{-1} (\mathbf{x} - \boldsymbol{\mu}_1) = (\mathbf{x} - \boldsymbol{\mu}_2)^\top \mathbf{\Sigma}_2^{-1} (\mathbf{x} - \boldsymbol{\mu}_2)
$$
An unnormalized Gaussian distribution is uniquely defined by its mean and covariance. Therefore, this equality implies $\boldsymbol{\mu}_1 = \boldsymbol{\mu}_2$ and $\mathbf{\Sigma}_1 = \mathbf{\Sigma}_2$.

The submanifold $\mathcal{M}_i$ is defined as the level set:
$$
\mathcal{M}_i = \{ \mathbf{x} \in \mathbb{R}^3 \mid (\mathbf{x} - \boldsymbol{\mu}_i)^\top \mathbf{\Sigma}_i^{-1} (\mathbf{x} - \boldsymbol{\mu}_i) = r^2 \}
$$
Since the parameters $(\boldsymbol{\mu}_i, \mathbf{\Sigma}_i)$ that define the level set are identical for both primitives, the resulting sets of points must also be identical. Thus, $\mathcal{M}_1 = \mathcal{M}_2$.

The submanifold color field $F_i$ is defined for a point $\mathbf{x} \in \mathcal{M}_i$ as:
$$
F_i(\mathbf{x}) = c_{\mathcal{G}_i}(\mathbf{d}_{\mathbf{x}}), \quad \text{where} \quad \mathbf{d}_{\mathbf{x}} = (\mathbf{x} - \boldsymbol{\mu}_i) / \lVert\mathbf{x} - \boldsymbol{\mu}_i\rVert
$$
From the hypothesis, we know that $c_{\mathcal{G}_1}(\mathbf{d}) = c_{\mathcal{G}_2}(\mathbf{d})$ holds for any unit direction vector $\mathbf{d} \in \mathbb{S}^2$.

For any point $\mathbf{x}$ on the common manifold $\mathcal{M} = \mathcal{M}_1 = \mathcal{M}_2$, its corresponding direction vector $\mathbf{d}_{\mathbf{x}}$ is an element of $\mathbb{S}^2$. We can therefore apply the hypothesis for this specific direction:
$$
c_{\mathcal{G}_1}(\mathbf{d}_{\mathbf{x}}) = c_{\mathcal{G}_2}(\mathbf{d}_{\mathbf{x}})
$$
By the definition of $F_i$, this directly implies:
$$
F_1(\mathbf{x}) = F_2(\mathbf{x})
$$
This holds for all $\mathbf{x} \in \mathcal{M}$. Thus, the color fields $F_1$ and $F_2$ are identical.

\section{Implementation details}
\label{app:implementation}

\subsection{Submanifold Field VAE}

For completeness, we note a few aspects not detailed in the main text. Also see Alg. \ref{alg:sfvae} for steps in one training step.

\textbf{Uniform Point Sampling.} The submanifold field $\mathcal{E}_i$ is discretized by using a uniform mesh grid of size $(n, n)$ to sample $P=n^2$ points on the ellipsoidal surface $\mathcal{M}_i$ with respect to area, forming $\mathcal{P}_i = \{ (\mathbf{x}_{i,k}, F_i(\mathbf{x}_{i,k}), \alpha_i) \}_{k=1}^P$.

\textbf{Decoding Gaussian Parameters.} After decoding, we recover Gaussian parameters from the reconstructed point cloud by first applying batched PCA to estimate the ellipsoid axes and orientation: we compute the mean and covariance of the points, perform eigen decomposition to obtain principal axes, and ensure a right-handed coordinate system. The logarithm of the axis lengths gives the scale parameters, and the rotation matrix is converted to a quaternion using a numerically stable batched algorithm. For appearance, we compute ellipsoid-normalized directions for each point and fit spherical harmonics coefficients to the RGB values via regularized batched least-squares. Opacity is estimated by averaging and logit-transforming the per-point values.

\begin{algorithm}[H]
\caption{SF-VAE: one training step for a minibatch of submanifold fields}
\label{alg:sfvae}
\begin{algorithmic}[1]
\Require Batch $\{(\bm{\mu}_i,\bm{\Sigma}_i,\mathbf{c}_i,o_i)\}_{i=1}^B$, fixed $r^2{=}1$, point count $P$
\For{each $i$}
    \State \textit{(Sampling on $\mathcal{M}_i$)} Sample $\{\mathbf{u}_k\}_{k=1}^{P}$ quasi-uniformly on $\mathbb{S}^2$; set $\mathbf{x}_{i,k}\gets\bm{\mu}_i + \bm{\Sigma}_i^{1/2} \mathbf{u}_k$ so that $(\mathbf{x}_{i,k}-\bm{\mu}_i)^\top\bm{\Sigma}_i^{-1}(\mathbf{x}_{i,k}-\bm{\mu}_i)=r^2$
    \State \textit{(Color/opacity)} $\mathbf{d}_{i,k}\gets(\mathbf{x}_{i,k}-\bm{\mu}_i)/\|\mathbf{x}_{i,k}-\bm{\mu}_i\|$;\;\; $\mathbf{c}_{i,k}\gets\mathrm{SH}(\mathbf{c}_i,\mathbf{d}_{i,k})$;\;\; $\alpha_{i,k}\gets\sigma(o_i)$
    \State \textit{(Point set)} $\mathcal{P}_i\gets\{(\mathbf{x}_{i,k},\mathbf{c}_{i,k},\alpha_{i,k})\}_{k=1}^{P}$
\EndFor
\State \textit{(Encode)} $(\bm{\mu}^z_i,\log\bm{\sigma}^{2,z}_i)\gets E(\mathcal{P}_i)$;\; $\mathbf{z}_i\gets\bm{\mu}^z_i+\bm{\sigma}^z_i\odot\varepsilon,\;\varepsilon\sim\mathcal{N}(0,I)$
\State \textit{(Decode)} $(\widehat{\mathcal{M}}_i,\widehat{F}_i,\hat\alpha_i)\gets D(\mathbf{z}_i)$;\; $\widehat{\mathcal{P}}_i\gets\{(\hat{\mathbf{x}}_{i,k},\widehat{F}_i(\hat{\mathbf{x}}_{i,k}),\hat\alpha_i)\}$
\State \textit{(Recover parameters)} $\widehat{\bm{\Sigma}}_i\gets\mathrm{PCA}(\{\hat{\mathbf{x}}_{i,k}\})$;\; $\widehat{\mathbf{c}}_i\gets\arg\min_{\mathbf{c}}\sum_k\|\mathrm{SH}(\mathbf{c},\hat{\mathbf{d}}_{i,k})-\widehat{F}_i(\hat{\mathbf{x}}_{i,k})\|_2^2$
\State \textit{(Loss)} $\mathcal{L}_{\text{rec}}\gets W_2^{(\varepsilon)}(\mathcal{P}_i,\widehat{\mathcal{P}}_i)$ with $d(\cdot,\cdot)$ from Eq.\,(8);\; $\mathcal{L}\gets\mathcal{L}_{\text{rec}}+\beta\cdot\mathrm{KL}$
\State \textit{(Update)} $\boldsymbol{\theta}_E,\boldsymbol{\theta}_D \leftarrow \boldsymbol{\theta}_E,\boldsymbol{\theta}_D - \eta \nabla_{\boldsymbol{\theta}_E,\boldsymbol{\theta}_D}\mathcal{L}$
\end{algorithmic}
\end{algorithm}

\subsection{Generated 3D Gaussian primitives dataset}

\textbf{Parameter priors and sampling}
Each primitive $\mathcal{G}_i$ is sampled as $\boldsymbol{\theta}_i=\{\boldsymbol{\mu}_i,\mathbf{q}_i,\mathbf{s}_i,\mathbf{c}_i,o_i\}$ and converted to $(\boldsymbol{\mu}_i,\Sigma_i,\mathbf{c}_i,\alpha_i)$ with
$\Sigma_i=R(\mathbf{q}_i)\,\mathrm{diag}(\exp(\mathbf{s}_i))^2\,R(\mathbf{q}_i)^\top$ and $\alpha_i=\sigma(o_i)$. Unless otherwise stated we use:
\begin{itemize}
\item \textbf{Mean.} $\boldsymbol{\mu}_i = (0, 0, 0)$. Since our setting only samples single Gaussians, extrinsic information is not needed.
\item \textbf{Rotation.} $\mathbf{q}_i$ is sampled uniformly on $SO(3)$ (normalized and, if needed, enforce a canonical sign).
\item \textbf{Scale.} Log–axes $\mathbf{s}_i\in\mathbb{R}^3$ drawn i.i.d.\ from $\mathcal{U}([s_{min},s_{max}])$; set activated scales $\exp(s_i)$.
\item \textbf{SH coefficients.} 
Let $\beta>1$ be the decay factor (in code, $\beta=4$). For degree $\ell=0,\dots,L$,
we draw the $(2\ell{+}1)$-dimensional SH band as
$
\mathbf{c}_\ell \sim \mathcal{N}\!\big( \mathbf{0},\, \sigma_\ell^2 I_{2\ell+1} \big),
\qquad
\sigma_\ell \;=\; \beta^{-\ell},
$
i.e.,
$
\mathrm{Var}[\,c_{\ell,m}\,] \;=\; \beta^{-2\ell}\quad\text{for } m=-\ell,\dots,\ell.
$
If coefficients above the chosen degree $L$ are padded up to $L_{\max}$,
we use i.i.d.\ noise $c_{\ell,m}\sim\mathcal{N}(0,\sigma_{\text{void}}^2)$ with $\sigma_{\text{void}}=0.05$.
\item \textbf{Opacity.} Logit $o_i \sim \mathcal{U}([o_{min},o_{max}])$, with $\alpha_i=\sigma(o_i)$.
\end{itemize}

To sum up, we can describe the data distribution in this dataset as:
\[
\begin{aligned}
\mathcal{D}
&= \bigl\{(\boldsymbol{\mu}_i,\mathbf{q}_i,\mathbf{s}_i,\mathbf{c}_i,o_i)\bigr\}_{i=1}^N
\stackrel{\text{i.i.d.}}{\sim}
\underbrace{\delta_{\mathbf{0}}}_{\boldsymbol{\mu}}
\times
\underbrace{\mathcal{U}\!\big(SO(3)\big)}_{\mathbf{q}}
\times
\underbrace{\mathcal{U}\!\big([s_{\min},s_{\max}]\big)^{3}}_{\mathbf{s}}
\\[2pt]
&\quad \times\;
\underbrace{\left(
\prod_{c\in\{R,G,B\}}
\Big[
\prod_{\ell=0}^{L}\mathcal{N}\!\big(0,\beta^{-2\ell}\big)^{2\ell+1}
\;\times\;
\prod_{\ell=L+1}^{L_{\max}}\mathcal{N}\!\big(0,\sigma_{\text{void}}^{2}\big)^{2\ell+1}
\Big]
\right)}_{\mathbf{c}\in\mathbb{R}^{3\times K},\;K=(L_{\max}+1)^2}
\\[2pt]
&\quad \times\;
\underbrace{\mathcal{U}\!\big([o_{\min},o_{\max}]\big)}_{o},
\qquad
\beta>1 \ (\text{default }4),\ \ \sigma_{\text{void}}=0.05.
\end{aligned}
\]

where $\beta>1$ is the SH variance–decay factor (default $\beta{=}4$) and $\sigma_{\text{void}}$ is the padding noise std (default $0.05$). Activations used downstream are
$\Sigma_i=R(\mathbf{q}_i)\,\mathrm{diag}\!\big(\exp(\mathbf{s}_i)\big)^{2}R(\mathbf{q}_i)^\top$
and $\alpha_i=\sigma(o_i)$.

\textbf{Defaults and ablations}
Default hyperparameters: $N{=}500$K, $L_{\max}{=}3$ ($K{=}16$), $P{=}144$, $s_{min}{=}-8$, $s_{max}{=}0$, $o_{min}{=}-5$, $o_{max}{=}10$. These parameters are built from statistical analysis of data distribution in diverse 3DGS datasets. We ablate $P\in\{36,64,144,576\}$ in the main paper to assess reconstruction vs.\ cost.

\textbf{Data Formatting and Processing.}
For each of the randomly synthesized primitive we store the native tuple $\boldsymbol{\theta}_i$ (float32 arrays for $\boldsymbol{\mu}$, $\mathbf{q}$, $\mathbf{s}$, $\mathbf{c}$, $o$). The discretized field $\mathcal{P}_i$ is sampled from the primitive of $\boldsymbol{\theta}_i$ to a tensor of $(B, P, 7)$ where $7$ is $(x, y, z, r, g, b, \alpha)$. It can be obtained at runtime to reduce memory and storage requirements. The dataset exposes toggles to return either $\boldsymbol{\theta}$ or $\mathcal{P}$.

\begin{algorithm}[t]
\caption{\textsc{GaussianGen}: Generate one colored point cloud from a random Gaussian}
\label{alg:gaussiangen}
\begin{algorithmic}[1]
\Require point count $P$ (default $144$), SH degree $L_{\max}$ with $K=(L_{\max}{+}1)^2$
\State \textbf{Sample raw parameters} $\bm{\mu} \in \mathbb{R}^3$, $\mathbf{s} \in \mathbb{R}^3$, $\mathbf{q} \in SO(3)$, $o \in \mathbb{R}$,
$\mathrm{feat\_dc} \in \mathbb{R}^3$, $\mathrm{feat\_extra} \in \mathbb{R}^{3(K-1)}$
\State \textbf{Assemble SH coefficients} $\mathbf{C} \in \mathbb{R}^{3 \times K}$ by stacking per channel:
\Statex \hspace{1.8em} $\mathbf{C}_r \leftarrow [\mathrm{feat\_dc}[0],\;\mathrm{feat\_extra}[0:(K-1)]]$
\Statex \hspace{1.8em} $\mathbf{C}_g \leftarrow [\mathrm{feat\_dc}[1],\;\mathrm{feat\_extra}[(K-1):2(K-1)]]$
\Statex \hspace{1.8em} $\mathbf{C}_b \leftarrow [\mathrm{feat\_dc}[2],\;\mathrm{feat\_extra}[2(K-1):3(K-1)]]$
\State \textbf{Activate parameters}
\Statex \hspace{1.8em} $\mathbf{q} \leftarrow \mathbf{q} / \|\mathbf{q}\|$;\; $R \leftarrow R(\mathbf{q})$;\; $\bm{\sigma} \leftarrow \exp(\mathbf{s})$;\; $\alpha \leftarrow \sigma(o)$
\State \textbf{Build surface grid}
\Statex \hspace{1.8em} $n \leftarrow \sqrt{P}$;\; create angular grids $u\!\in\![0,2\pi)$ and $v\!\in\![0,\pi]$ of size $n\times n$
\Statex \hspace{1.8em} directions $\mathbf{d}(u,v)\in\mathbb{S}^2$ from spherical angles $(u,v)$
\State \textbf{Map to ellipsoid (iso-density $r^2{=}1$)}
\Statex \hspace{1.8em} $\mathbf{x}(u,v) \leftarrow \bm{\mu} + R\,\mathrm{diag}(\bm{\sigma})\,\mathbf{d}(u,v)$
\State \textbf{Evaluate color field}
\Statex \hspace{1.8em} $\mathbf{c}(u,v) \leftarrow \mathrm{SH}(\mathbf{C},\,\mathbf{d}(u,v))$
\Statex \hspace{1.8em} \textit{(optional)} color post-process: $\mathbf{c} \leftarrow \mathrm{clip}(\mathbf{c}{+}0.5,\;0,1)$
\State \textbf{Pack output}
\Statex \hspace{1.8em} replicate $\alpha$ to all points;\; flatten grids to $P$ points
\Statex \hspace{1.8em} \textbf{return} $\{(\mathbf{x}_k,\;\mathbf{c}_k,\;\alpha)\}_{k=1}^{P}\in\mathbb{R}^{P\times 7}$ \; // $(x,y,z,r,g,b,\alpha)$
\end{algorithmic}
\end{algorithm}

\newpage
\section{Additional Evaluation and Visualization Results}
\label{app:extra_results}

\subsection{Interpolation \& Noise Visualizations}

\begin{figure}[h]
    \centering
    \includegraphics[width=\linewidth]{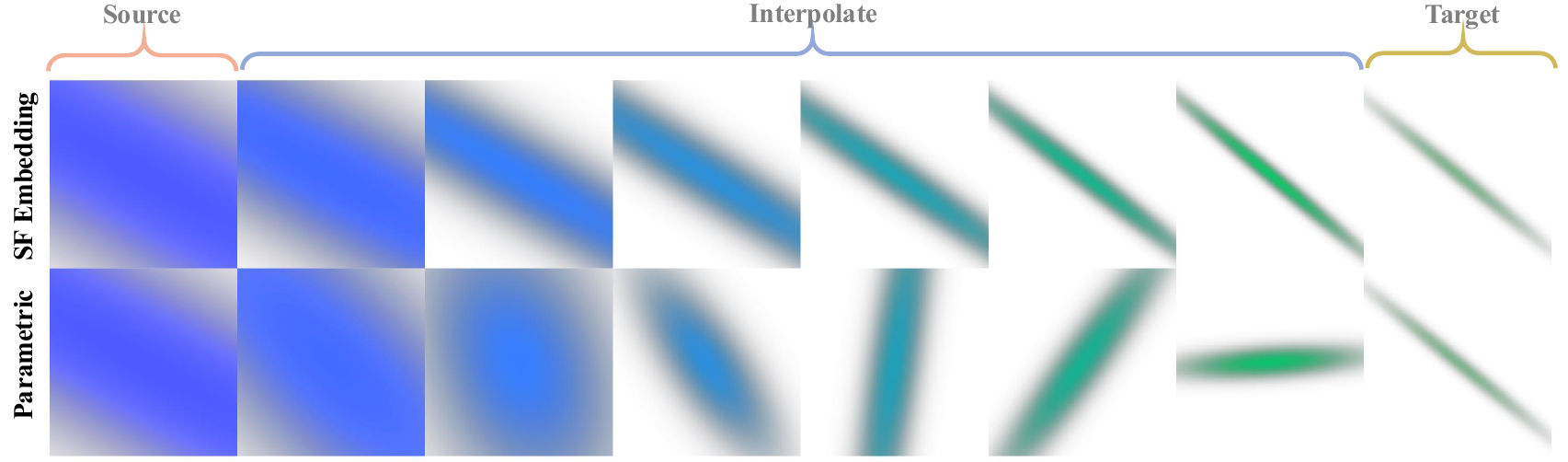}
    \caption{Sample visual comparison of linear interpolation in submanifold field embedding space and parametric space. Interpolation in SF embedding shows smooth transition from source to target. Perturb $\mathbf{s}$ and $\mathbf{q}$ will dramatically change the geometry of the resulted Gaussian, verifying the feature heterogeneous problem.}
    \label{fig:interpolate}
\end{figure}

\begin{figure}[h]
    \centering
    \includegraphics[width=0.8\linewidth]{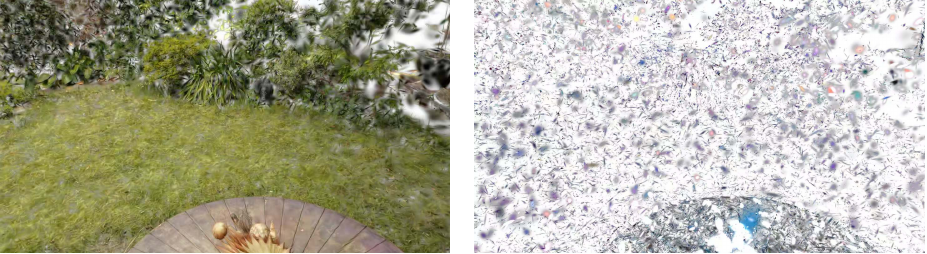}
    \caption{When rotation $\mathbf{q}$ is inverted to $-\mathbf{q}$, the submanifold field model (left) can still correctly reconstruct the scene, while parametric model fails to process this equivalent rotation. }
    \label{fig:perturb}
\end{figure}

\newpage
\subsection{Additional Reconstruction Visualizations}

\begin{figure}[h]
    \centering
    \includegraphics[width=0.8\linewidth]{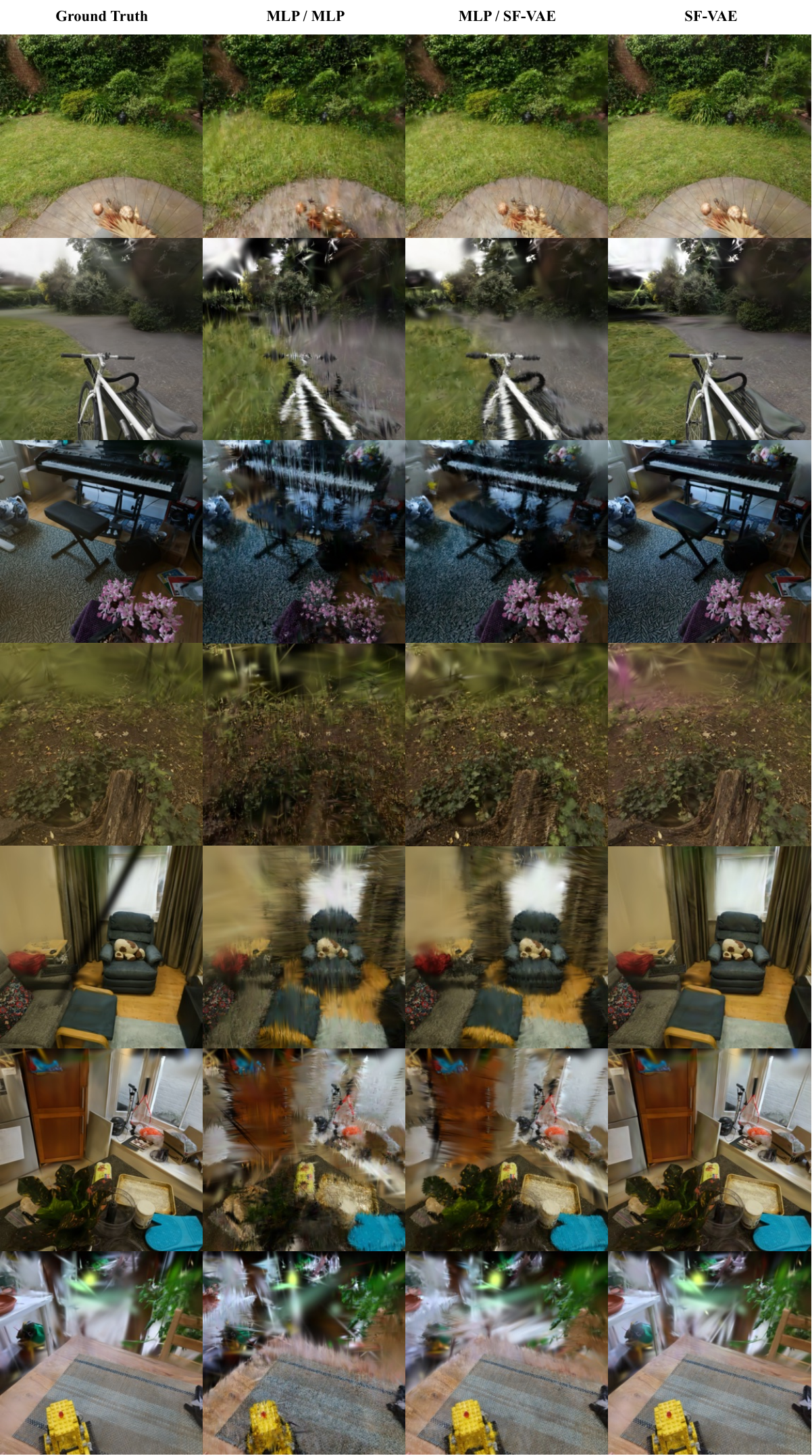}
    \caption{Full qualitative results for rasterization using  different model configurations on Mip-NeRF 360.}
    \label{fig:app_render}
\end{figure}

\newpage
\subsection{Unsupervised Clustering Quantitative Results}
\label{app:cluster_res}

\begin{table}[h]
    \centering
    \footnotesize
    \begin{tabular}{l|cc}
        \toprule
        \textbf{Clustered Feature} & \textbf{Silhouette} $\uparrow$ & \textbf{Cluster Compactness (MSE} $\downarrow$\textbf{)} \\
        \midrule
        Parameters & 0.090 & 71.306 \\
        MLP / MLP & 0.084 & 50.761 \\
        MLP / SF-VAE & 0.086 & 50.916 \\
        SF-VAE & \textbf{0.113} & \textbf{48.282} \\
        \bottomrule
    \end{tabular}
    \caption{Quantitative comparisons on unsupervised clustering based on the respective clustered feature. Higher silhouette is better; lower compactness (MSE) is better.}
    \label{tab:clustering_metrics}
\end{table}

\subsection{Gaussian Neural Field Details \& Visualizations}
\label{app:gnf}

\textbf{Network Structure for Gaussian Neural Fields.} 
We define a compact per-scene MLP that takes only 3D coordinates ((x,y,z)), applies sinusoidal/Fourier positional encoding, and passes the encoded vector through a fully connected backbone (256-width 4 layers for ShapeSplat, 512-width 8 layers for Mip-NeRF 360) with a single mid-network skip that re-concatenates the encoded input for better conditioning. The \emph{same network architecture} is used for both targets; the only difference is the final output dimension: either 32 for the SF embedding or 56 for the no-position Gaussian parameters ($[q(4),s(3),\text{SH}(3K),o(1)]$). Hidden layers use standard ReLU/SiLU activations, the output is linear, and the overall footprint remains lightweight ($\approx 2 \times 10^5$ parameters for ShapeSplat, $\approx 1.8\times 10^6$ for Mip-NeRF 360) for querying at Gaussian centers to yield SF embeddings or directly renderable Gaussian parameters.

\begin{figure}[h]
    \centering
    \includegraphics[width=0.6\linewidth]{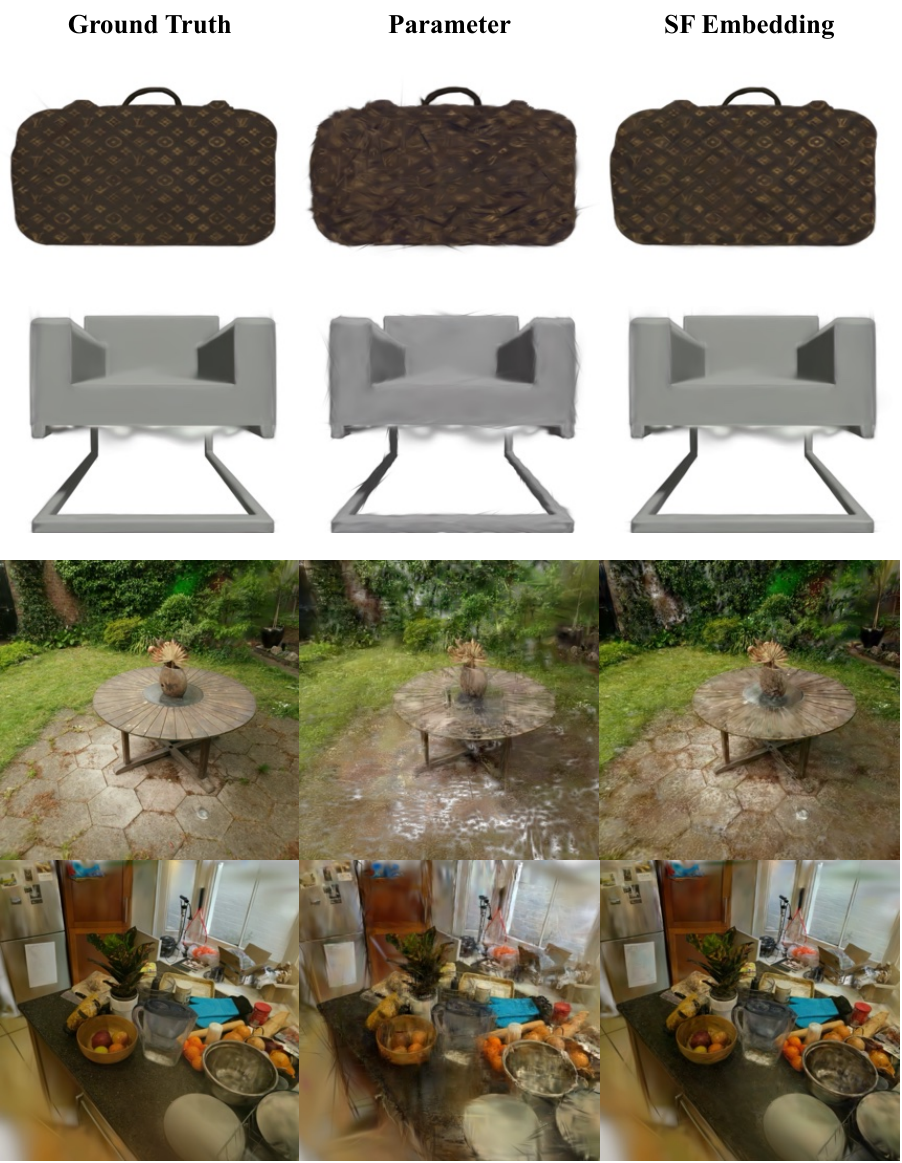}
    \caption{Arbitrarily picked visual comparisons of Gaussian Neural Fields trained with target Gaussian parameters and submanifold field embeddings. The submanifold field embedding results is qualitatively better under matching settings, indicating that the SF embedding is a better learning target for regression tasks such as training a neural field.}
    \label{fig:app_genf}
\end{figure}

%% file: iclr2026_conference.bib
@String(CVPR  = {CVPR})

@String(ICCV  = {ICCV})

@String(ECCV  = {ECCV})

@String(NIPS  = {NeurIPS})

@String(TOG   = {ACM TOG})

@String(AAAI = {AAAI})

@String(SIGGRAPH = {SIGGRAPH})

@String(ICML = {ICML})

@article{shapenet2015,
  title={Shapenet: An information-rich 3d model repository},
  author={Chang, Angel X and Funkhouser, Thomas and Guibas, Leonidas and Hanrahan, Pat and Huang, Qixing and Li, Zimo and Savarese, Silvio and Savva, Manolis and Song, Shuran and Su, Hao and others},
  journal={arXiv preprint arXiv:1512.03012},
  year={2015}
}

@inproceedings{barron2022mipnerf360unboundedantialiased,
  title={Mip-nerf 360: Unbounded anti-aliased neural radiance fields},
  author={Barron, Jonathan T and Mildenhall, Ben and Verbin, Dor and Srinivasan, Pratul P and Hedman, Peter},
  booktitle=CVPR,
  year={2022}
}

@inproceedings{qi2017pointnet,
  title={Pointnet: Deep learning on point sets for 3d classification and segmentation},
  author={Qi, Charles R and Su, Hao and Mo, Kaichun and Guibas, Leonidas J},
  booktitle={Proceedings of the IEEE conference on computer vision and pattern recognition},
  pages={652--660},
  year={2017}
}

@article{bengio2013representation,
  title={Representation learning: A review and new perspectives},
  author={Bengio, Yoshua and Courville, Aaron and Vincent, Pascal},
  journal={IEEE transactions on pattern analysis and machine intelligence},
  volume={35},
  number={8},
  pages={1798--1828},
  year={2013},
  publisher={IEEE}
}

@inproceedings{wang2020understanding,
  title={Understanding contrastive representation learning through alignment and uniformity on the hypersphere},
  author={Wang, Tongzhou and Isola, Phillip},
  booktitle={International conference on machine learning},
  pages={9929--9939},
  year={2020},
  organization={PMLR}
}

@inproceedings{ioffe2015batch,
  title={Batch normalization: Accelerating deep network training by reducing internal covariate shift},
  author={Ioffe, Sergey and Szegedy, Christian},
  booktitle=ICML,
  year={2015}
}

@article{kerbl20233d,
  title={3D Gaussian splatting for real-time radiance field rendering.},
  author={Kerbl, Bernhard and Kopanas, Georgios and Leimk{\"u}hler, Thomas and Drettakis, George},
  journal={ACM Trans. Graph.},
  year={2023}
}

@article{bao20253d,
  title={3d gaussian splatting: Survey, technologies, challenges, and opportunities},
  author={Bao, Yanqi and Ding, Tianyu and Huo, Jing and Liu, Yaoli and Li, Yuxin and Li, Wenbin and Gao, Yang and Luo, Jiebo},
  journal={IEEE Transactions on Circuits and Systems for Video Technology},
  year={2025},
  publisher={IEEE}
}

@article{jo2024identifying,
  title={Identifying unnecessary 3d gaussians using clustering for fast rendering of 3d gaussian splatting},
  author={Jo, Joongho and Kim, Hyeongwon and Park, Jongsun},
  journal={arXiv preprint arXiv:2402.13827},
  year={2024}
}

@inproceedings{lee2024gscore,
  title={Gscore: Efficient radiance field rendering via architectural support for 3d gaussian splatting},
  author={Lee, Junseo and Lee, Seokwon and Lee, Jungi and Park, Junyong and Sim, Jaewoong},
  booktitle={Proceedings of the 29th ACM International Conference on Architectural Support for Programming Languages and Operating Systems, Volume 3},
  pages={497--511},
  year={2024}
}

@inproceedings{ma2025large,
  title={A Large-Scale Dataset of Gaussian Splats and Their Self-Supervised Pretraining},
  author={Ma, Qi and Li, Yue and Ren, Bin and Sebe, Nicu and Konukoglu, Ender and Gevers, Theo and Van Gool, Luc and Paudel, Danda Pani},
  booktitle={2025 International Conference on 3D Vision (3DV)},
  year={2025},
}

@article{li2025scenesplat,
  title={Scenesplat: Gaussian splatting-based scene understanding with vision-language pretraining},
  author={Li, Yue and Ma, Qi and Yang, Runyi and Li, Huapeng and Ma, Mengjiao and Ren, Bin and Popovic, Nikola and Sebe, Nicu and Konukoglu, Ender and Gevers, Theo and others},
  journal=ICCV,
  year={2025}
}

@inproceedings{kocabas2024hugs,
  title={Hugs: Human gaussian splats},
  author={Kocabas, Muhammed and Chang, Jen-Hao Rick and Gabriel, James and Tuzel, Oncel and Ranjan, Anurag},
  booktitle=CVPR,
  year={2024}
}

@inproceedings{li2024animatable,
  title={Animatable gaussians: Learning pose-dependent gaussian maps for high-fidelity human avatar modeling},
  author={Li, Zhe and Zheng, Zerong and Wang, Lizhen and Liu, Yebin},
  booktitle=CVPR,
  year={2024}
}

@inproceedings{zhou2024drivinggaussian,
  title={Drivinggaussian: Composite gaussian splatting for surrounding dynamic autonomous driving scenes},
  author={Zhou, Xiaoyu and Lin, Zhiwei and Shan, Xiaojun and Wang, Yongtao and Sun, Deqing and Yang, Ming-Hsuan},
  booktitle=CVPR,
  year={2024}
}

@article{yan2023street,
  title={Street Gaussians for Modeling Dynamic Urban Scenes.(2023)},
  author={Yan, Yunzhi and Lin, Haotong and Zhou, Chenxu and Wang, Weijie and Sun, Haiyang and Zhan, Kun and Lang, Xianpeng and Zhou, Xiaowei and Peng, Sida},
  journal={arXiv preprint arXiv:2401.01339},
  year={2023}
}

@inproceedings{zhou2024hugs,
  title={Hugs: Holistic urban 3d scene understanding via gaussian splatting},
  author={Zhou, Hongyu and Shao, Jiahao and Xu, Lu and Bai, Dongfeng and Qiu, Weichao and Liu, Bingbing and Wang, Yue and Geiger, Andreas and Liao, Yiyi},
  booktitle=CVPR,
  year={2024}
}

@article{xiaoyuan,
  title={HoliGS: Holistic Gaussian Splatting for Embodied View Synthesis},
  author={Wang, Xiaoyuan and Zhao, Yizhou and Ye, Botao and Shan, Xiaojun and Lyu, Weijie and Qi, Lu and Chan, Kelvin CK and Li, Yinxiao and Yang, Ming-Hsuan},
  journal=NIPS,
  year={2025}
}

@inproceedings{xinlin,
  title={HQGS: High-Quality Novel View Synthesis with Gaussian Splatting in Degraded Scenes},
  author={Lin, Xin and Luo, Shi and Shan, Xiaojun and Zhou, Xiaoyu and Ren, Chao and Qi, Lu and Yang, Ming-Hsuan and Vasconcelos, Nuno},
  booktitle={The Thirteenth International Conference on Learning Representations},
  year={2025}
}

@article{zhang2024stylizedgs,
  title={Stylizedgs: Controllable stylization for 3d gaussian splatting},
  author={Zhang, Dingxi and Yuan, Yu-Jie and Chen, Zhuoxun and Zhang, Fang-Lue and He, Zhenliang and Shan, Shiguang and Gao, Lin},
  journal={arXiv preprint arXiv:2404.05220},
  year={2024}
}

@article{sun2025generalizable,
  title={Generalizable and Relightable Gaussian Splatting for Human Novel View Synthesis},
  author={Sun, Yipengjing and Wang, Chenyang and Zheng, Shunyuan and Li, Zonglin and Zhang, Shengping and Ji, Xiangyang},
  journal={arXiv preprint arXiv:2505.21502},
  year={2025}
}

@inproceedings{lee2025editsplat,
  title={Editsplat: Multi-view fusion and attention-guided optimization for view-consistent 3d scene editing with 3d gaussian splatting},
  author={Lee, Dong In and Park, Hyeongcheol and Seo, Jiyoung and Park, Eunbyung and Park, Hyunje and Baek, Ha Dam and Shin, Sangheon and Kim, Sangmin and Kim, Sangpil},
  booktitle=CVPR,
  year={2025}
}

@inproceedings{charatan2024pixelsplat,
  title={pixelsplat: 3d gaussian splats from image pairs for scalable generalizable 3d reconstruction},
  author={Charatan, David and Li, Sizhe Lester and Tagliasacchi, Andrea and Sitzmann, Vincent},
  booktitle=CVPR,
  year={2024}
}

@inproceedings{chen2024mvsplat,
  title={Mvsplat: Efficient 3d gaussian splatting from sparse multi-view images},
  author={Chen, Yuedong and Xu, Haofei and Zheng, Chuanxia and Zhuang, Bohan and Pollefeys, Marc and Geiger, Andreas and Cham, Tat-Jen and Cai, Jianfei},
  booktitle=ECCV,
  year={2024},
}

@inproceedings{zheng2024gps,
  title={Gps-gaussian: Generalizable pixel-wise 3d gaussian splatting for real-time human novel view synthesis},
  author={Zheng, Shunyuan and Zhou, Boyao and Shao, Ruizhi and Liu, Boning and Zhang, Shengping and Nie, Liqiang and Liu, Yebin},
  booktitle=CVPR,
  year={2024}
}

@article{hong2024pf3plat,
  title={Pf3plat: Pose-free feed-forward 3d gaussian splatting},
  author={Hong, Sunghwan and Jung, Jaewoo and Shin, Heeseong and Han, Jisang and Yang, Jiaolong and Luo, Chong and Kim, Seungryong},
  journal=ICML,
  year={2024}
}

@article{chen2024pref3r,
  title={Pref3r: Pose-free feed-forward 3d gaussian splatting from variable-length image sequence},
  author={Chen, Zequn and Yang, Jiezhi and Yang, Heng},
  journal={arXiv preprint arXiv:2411.16877},
  year={2024}
}

@inproceedings{tian2025drivingforward,
  title={Drivingforward: Feed-forward 3d gaussian splatting for driving scene reconstruction from flexible surround-view input},
  author={Tian, Qijian and Tan, Xin and Xie, Yuan and Ma, Lizhuang},
  booktitle=AAAI,
  year={2025}
}

@article{li2025vicasplat,
  title={Vicasplat: A single run is all you need for 3d gaussian splatting and camera estimation from unposed video frames},
  author={Li, Zhiqi and Dong, Chengrui and Chen, Yiming and Huang, Zhangchi and Liu, Peidong},
  journal={arXiv preprint arXiv:2503.10286},
  year={2025}
}

@article{jiang2025anysplat,
  title={AnySplat: Feed-forward 3D Gaussian Splatting from Unconstrained Views},
  author={Jiang, Lihan and Mao, Yucheng and Xu, Linning and Lu, Tao and Ren, Kerui and Jin, Yichen and Xu, Xudong and Yu, Mulin and Pang, Jiangmiao and Zhao, Feng and others},
  journal={arXiv preprint arXiv:2505.23716},
  year={2025}
}

@article{lin2025longsplat,
  title={LongSplat: Robust Unposed 3D Gaussian Splatting for Casual Long Videos},
  author={Lin, Chin-Yang and Sun, Cheng and Yang, Fu-En and Chen, Min-Hung and Lin, Yen-Yu and Liu, Yu-Lun},
  journal=ICCV,
  year={2025}
}

@inproceedings{girish2024eagles,
  title={Eagles: Efficient accelerated 3d gaussians with lightweight encodings},
  author={Girish, Sharath and Gupta, Kamal and Shrivastava, Abhinav},
  booktitle=ECCV,
  year={2024}
}

@inproceedings{qin2024langsplat,
  title={Langsplat: 3d language gaussian splatting},
  author={Qin, Minghan and Li, Wanhua and Zhou, Jiawei and Wang, Haoqian and Pfister, Hanspeter},
  booktitle=CVPR,
  year={2024}
}

@article{zhobro2025learning,
  title={Learning 3D-Gaussian Simulators from RGB Videos},
  author={Zhobro, Mikel and Geist, Andreas Ren{\'e} and Martius, Georg},
  journal={arXiv preprint arXiv:2503.24009},
  year={2025}
}

@article{zhou2024diffgs,
  title={Diffgs: Functional gaussian splatting diffusion},
  author={Zhou, Junsheng and Zhang, Weiqi and Liu, Yu-Shen},
  journal=NIPS,
  year={2024}
}

@article{lin2025diffsplat,
  title={Diffsplat: Repurposing image diffusion models for scalable gaussian splat generation},
  author={Lin, Chenguo and Pan, Panwang and Yang, Bangbang and Li, Zeming and Mu, Yadong},
  journal={arXiv preprint arXiv:2501.16764},
  year={2025}
}

@inproceedings{wewer2024latentsplat,
  title={latentsplat: Autoencoding variational gaussians for fast generalizable 3d reconstruction},
  author={Wewer, Christopher and Raj, Kevin and Ilg, Eddy and Schiele, Bernt and Lenssen, Jan Eric},
  booktitle=ECCV,
  year={2024}
}

@inproceedings{chen2024gaussianeditor,
  title={Gaussianeditor: Swift and controllable 3d editing with gaussian splatting},
  author={Chen, Yiwen and Chen, Zilong and Zhang, Chi and Wang, Feng and Yang, Xiaofeng and Wang, Yikai and Cai, Zhongang and Yang, Lei and Liu, Huaping and Lin, Guosheng},
  booktitle=CVPR,
  year={2024}
}

@misc{igs2gs,
         author = {Vachha, Cyrus and Haque, Ayaan},
         title = {Instruct-GS2GS: Editing 3D Gaussian Splats with Instructions},
         year = {2024},
         url = {https://instruct-gs2gs.github.io/}
        }

@article{palandra2024gsedit,
  title={Gsedit: Efficient text-guided editing of 3d objects via gaussian splatting},
  author={Palandra, Francesco and Sanchietti, Andrea and Baieri, Daniele and Rodola, Emanuele},
  journal={arXiv preprint arXiv:2403.05154},
  year={2024}
}

@inproceedings{kovacs2024,
  title={G-Style: Stylized Gaussian Splatting},
  author={Kov{\'a}cs, {\'A}ron Samuel and Hermosilla, Pedro and Raidou, Renata G},
  booktitle={Computer Graphics Forum},
  year={2024},
}

@article{yu2024instantstylegaussian,
  title={Instantstylegaussian: Efficient art style transfer with 3d gaussian splatting},
  author={Yu, Xin-Yi and Yu, Jun-Xin and Zhou, Li-Bo and Wei, Yan and Ou, Lin-Lin},
  journal={arXiv preprint arXiv:2408.04249},
  year={2024}
}

@article{tang2023dreamgaussian,
  title={Dreamgaussian: Generative gaussian splatting for efficient 3d content creation},
  author={Tang, Jiaxiang and Ren, Jiawei and Zhou, Hang and Liu, Ziwei and Zeng, Gang},
  journal={arXiv preprint arXiv:2309.16653},
  year={2023}
}

@inproceedings{yi2024gaussiandreamer,
  title={Gaussiandreamer: Fast generation from text to 3d gaussians by bridging 2d and 3d diffusion models},
  author={Yi, Taoran and Fang, Jiemin and Wang, Junjie and Wu, Guanjun and Xie, Lingxi and Zhang, Xiaopeng and Liu, Wenyu and Tian, Qi and Wang, Xinggang},
  booktitle=CVPR,
  year={2024}
}

@inproceedings{chen2024text,
  title={Text-to-3d using gaussian splatting},
  author={Chen, Zilong and Wang, Feng and Wang, Yikai and Liu, Huaping},
  booktitle=CVPR,
  year={2024}
}

@inproceedings{jiang2024vr,
  title={Vr-gs: A physical dynamics-aware interactive gaussian splatting system in virtual reality},
  author={Jiang, Ying and Yu, Chang and Xie, Tianyi and Li, Xuan and Feng, Yutao and Wang, Huamin and Li, Minchen and Lau, Henry and Gao, Feng and Yang, Yin and others},
  booktitle=SIGGRAPH,
  year={2024}
}

@inproceedings{xie2024physgaussian,
  title={Physgaussian: Physics-integrated 3d gaussians for generative dynamics},
  author={Xie, Tianyi and Zong, Zeshun and Qiu, Yuxing and Li, Xuan and Feng, Yutao and Yang, Yin and Jiang, Chenfanfu},
  booktitle=CVPR,
  year={2024}
}

@inproceedings{zhong2024reconstruction,
  title={Reconstruction and simulation of elastic objects with spring-mass 3d gaussians},
  author={Zhong, Licheng and Yu, Hong-Xing and Wu, Jiajun and Li, Yunzhu},
  booktitle=ECCV,
  year={2024}
}

@article{yang2025hybridgs,
  title={HybridGS: High-Efficiency Gaussian Splatting Data Compression using Dual-Channel Sparse Representation and Point Cloud Encoder},
  author={Yang, Qi and Yang, Le and Van Der Auwera, Geert and Li, Zhu},
  journal={arXiv preprint arXiv:2505.01938},
  year={2025}
}

@article{shin2025locality,
  title={Locality-aware gaussian compression for fast and high-quality rendering},
  author={Shin, Seungjoo and Park, Jaesik and Cho, Sunghyun},
  journal={arXiv preprint arXiv:2501.05757},
  year={2025}
}

@article{wu2024implicit,
  title={Implicit Gaussian Splatting with Efficient Multi-Level Tri-Plane Representation},
  author={Wu, Minye and Tuytelaars, Tinne},
  journal={CoRR},
  year={2024}
}

@inproceedings{xie2025generative,
  title={Generative Gaussian splatting for unbounded 3D city generation},
  author={Xie, Haozhe and Chen, Zhaoxi and Hong, Fangzhou and Liu, Ziwei},
  booktitle={Proceedings of the Computer Vision and Pattern Recognition Conference},
  pages={6111--6120},
  year={2025}
}

@article{guo2024semantic,
  title={Semantic gaussians: Open-vocabulary scene understanding with 3d gaussian splatting},
  author={Guo, Jun and Ma, Xiaojian and Fan, Yue and Liu, Huaping and Li, Qing},
  journal={arXiv preprint arXiv:2403.15624},
  year={2024}
}

@article{dgcnn,
  title={Dynamic Graph CNN for Learning on Point Clouds},
  author={Wang, Yue and Sun, Yongbin and Liu, Ziwei and Sarma, Sanjay E. and Bronstein, Michael M. and Solomon, Justin M.},
  journal={ACM Transactions on Graphics (TOG)},
  year={2019}
}
